\documentclass{article}

\PassOptionsToPackage{numbers, compress}{natbib}

\usepackage[preprint]{neurips_2024}

\usepackage[utf8]{inputenc} 
\usepackage[T1]{fontenc}    
\usepackage{hyperref}       
\usepackage{url}            
\usepackage{booktabs}       
\usepackage{amsfonts}       
\usepackage{nicefrac}       
\usepackage{microtype}      
\usepackage{xcolor}         
\usepackage{lineno,lettrine}
\usepackage{graphicx,subcaption,comment, makecell, colortbl}
\usepackage{amsmath,amssymb,multirow,bm} 
\usepackage{float,tikz,pgfplots, enumitem}
\usetikzlibrary{pgfplots.groupplots}
\pgfplotsset{compat=1.17}
\usepackage[normalem]{ulem}
\usepackage{pifont}
\newcommand{\cmark}{\ding{51}}%
\newcommand{\xmark}{\ding{53}}%

\title{SATA: Spatial Autocorrelation Token Analysis for Enhancing the Robustness of Vision Transformers}

\author{%
  Nick Nikzad\thanks{Corresponding Author} \\
  Griffith University\\
  QLD, Australia \\
  \texttt{n.nikzaddehaji@griffith.edu.au} \\
  \And
  Yi Liao  \\
  Griffith University \\
  QLD, Australia \\
  \texttt{yi.liao2@griffithuni.edu.au} \\
  \AND
  Yongsheng Gao \\
  Griffith University \\
  QLD, Australia \\
  \texttt{yongsheng.gao@griffith.edu.au} \\
  \And
  Jun Zhou \\
  Griffith University \\
  QLD, Australia \\
  \texttt{jun.zhou@griffith.edu.au} \\
}

\begin{document}

\maketitle

\begin{abstract}
 Over the past few years, vision transformers (ViTs) have consistently demonstrated remarkable performance across various visual recognition tasks. However, attempts to enhance their robustness have yielded limited success, mainly focusing on different training strategies, input patch augmentation, or network structural enhancements. These approaches often involve extensive training and fine-tuning, which are time-consuming and resource-intensive. To tackle these obstacles, we introduce a novel approach named Spatial Autocorrelation Token Analysis (SATA). By harnessing spatial relationships between token features, SATA enhances both the representational capacity and robustness of ViT models. This is achieved through the analysis and grouping of tokens according to their spatial autocorrelation scores prior to their input into the Feed-Forward Network (FFN) block of the self-attention mechanism. Importantly, SATA seamlessly integrates into existing pre-trained ViT baselines without requiring retraining or additional fine-tuning, while concurrently improving efficiency by reducing the computational load of the FFN units. Experimental results show that the baseline ViTs enhanced with SATA not only achieve a new state-of-the-art top-1 accuracy on ImageNet-1K image classification (94.9\%) but also establish new state-of-the-art performance across multiple robustness benchmarks, including ImageNet-A (top-1=63.6\%), ImageNet-R (top-1=79.2\%), and ImageNet-C (mCE=13.6\%), all without requiring additional training or fine-tuning of baseline models.
\end{abstract}

\section{Introduction}
\begin{figure}[t]
    \centering
    \includegraphics[trim=0cm 0.7cm 0.0cm 0.0cm, clip, scale=0.75]{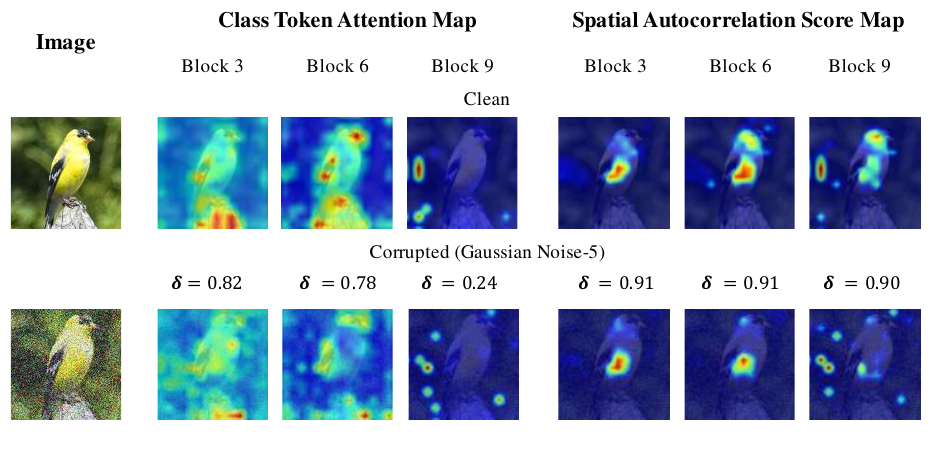}
    \caption{Visual comparison of class token attention maps and spatial autocorrelation score maps across three layers of Deit-Base/16 pre-trained on ImageNet-1K. The clean image is sourced from the 'goldfinch' class of the ImageNet-1K dataset, and its corresponding corrupted version, with maximum severity (5), is sourced from the ImageNet-C~\cite{ImageNet-C} dataset. $\delta$ represents the cosine similarity between either the attention maps or the spatial autocorrelation score maps of a corrupted image and its corresponding clean version at each block.}
    \label{fig:spa-score-map}
    \vspace*{-5mm}
\end{figure}
In recent years, vision transformers (ViTs) have demonstrated exceptional performance across diverse computer vision applications~\cite{vit,vit_survey_pami}. Drawing inspiration from the significant achievements of transformer architectures in natural language processing (NLP), ViTs divide an input image into a sequence of patches (tokens) and leverage self-attention layers~\cite{AttAll} to capture relationships between these tokens, ultimately generating rich representations for visual recognition tasks. While recent studies indicate that ViTs can exhibit greater robustness than Convolutional Networks (ConvNets), attributed to their self-attention mechanism~\cite{transformers_robust,vit_robust, intriguing_vit, understanding_robust_vit_cls}. However, this hypothesis has been challenged. Liu \textit{et al.}~\cite{ConvNeXt} demonstrated that a carefully constructed ConvNet can surpass ViTs in both generalization and robustness. Furthermore, while techniques such as patch augmentation\cite{understanding_vit_robust_aug, RVT}, contrastive learning strategies~\cite{understanding_vit_robust_aug, improving_robust_vit}, and network adjustments~\cite{RVT, FAN} have shown promise in enhancing ViT performance and robustness, they suffer from a significant drawback: they necessitate extensive retraining or fine-tuning on expansive datasets (\textit{e.g.}, ImageNet-1K, ImageNet-21K). This laborious and resource-intensive process poses a substantial bottleneck, particularly with large-scale ViT architectures.

Recently, Nikzad \textit{et al.}~\cite{csa_net} showed the existence of spatial correlation among feature maps in convolutional neural networks (CNNs). Moreover, they observed a decrease in spatial autocorrelation among feature maps through deeper network layers, suggesting that final features exhibit reduced spatial dependency. Motivated by these findings, in this work, we first investigate spatial autocorrelation within Vision Transformer (ViT) architectures and its implications for their performance and robustness. Then, we present a novel approach named ``\textit{Spatial Autocorrelation Token Analysis}'' (SATA) to tackle the identified shortcomings in the current efforts to enhance ViT robustness. In particular, our analysis confirms the presence of spatial autocorrelation among visual patches (tokens) and reveals a similar trend of decreasing overall spatial autocorrelation scores through ViT networks as observed in CNNs~\cite{csa_net}. Moreover, as illustrated in Figure~\ref{fig:spa-score-map}, our study shows that in the later layers of ViT networks, the spatial autocorrelation scores of patches are more robust against different corruptions compared to their attention maps. Additionally, patches with extremely high or low spatial autocorrelation scores in non-informative regions can impede recognition performance and compromise the network's robustness against corrupted inputs. To address this issue, our proposed SATA method adopts a unique splitting and grouping algorithm based on tokens' spatial relation scores in the later layers. This approach prevents the input of unnecessary tokens into the FFN block of the self-attention mechanism. Notably, SATA seamlessly integrates into various pre-trained ViT baselines without necessitating retraining or additional fine-tuning. This enhances ViT robustness and improves inference efficiency by reducing the computational load on the FFN units.

Extensive experiments conducted on ImageNet-1K image classification and various robustness evaluation benchmarks demonstrate the effectiveness of the proposed spatial autocorrelation paradigm in significantly improving the robustness and accuracy performance of Vision Transformers (ViTs). These findings establish a new state-of-the-art performance level, achieving a top-1 accuracy of 94.9\% on ImageNet-1K image classification, as well as impressive results across multiple robustness benchmarks, including ImageNet-A~\cite{ImageNet-A} (top-1=63.6\%), ImageNet-R~\cite{ImageNet-R} (top-1=79.2\%), and ImageNet-C~\cite{ImageNet-C} (mCE=13.6\%), without requiring additional expensive fine-tuning or training. Furthermore, in-depth investigations are conducted to thoroughly explore the characteristics of the proposed Spatial Autocorrelation Token Analysis (SATA).

\section{Related Works}
\paragraph{Vision Transformer.} Since the introduction of Vision Transformers (ViTs), they have achieved remarkable success in various computer vision tasks \cite{vit,vit_survey_pami}. Most improvements to date have focused on enhancing either the accuracy or the efficiency of ViTs. Numerous ViT variants have been proposed to boost their performance~\cite{vit_survey_pami}. Through dedicated data augmentation \cite{DeiT} and advanced self-attention structures \cite{yang2021focal, LeViT}, ViTs have demonstrated competitive or superior performance compared to convolutional neural networks (CNNs). Hybrid models like CvT~\cite{CvT} introduce intrinsic inductive bias into the ViT architecture by adding additional convolutional layers before the multi-head self-attention (MHSA) modules. CeiT~\cite{CeiT} extracts low-level features through the Image-to-Token (I2T) module and enhances locality by replacing the standard feed-forward network with the locally enhanced feed-forward (LeFF) layer. To enable ViTs to learn multi-scale features, CrossViT~\cite{CrossViT} employs a dual-branch transformer that combines different sizes of image patches to produce stronger image features. ViTAE~\cite{ViTAE} incorporates multi-scale context by designing reduction cells (RC) and normal cells (NC). 

To create efficient Vision Transformers, several recent works have focused on pruning~\cite{Adavit, Cp-vit, fayyaz2022adaptive} or combining~\cite{Spvit, liang2022not} tokens. ResT~\cite{ResT} introduces an efficient self-attention module using overlapping depth-wise convolutions, while T2T-ViT~\cite{T2T-ViT} employs a Tokens-to-Token (T2T) module for token aggregation. PiT~\cite{PiT} reduces spatial size with pooling layers, and Dynamic-ViT~\cite{DynamicViT} dynamically prunes tokens during inference. CaiT~\cite{CaiT} optimizes the ViT architecture with layer scaling and class-attention mechanisms. More recently, Bolya \textit{et al.}\cite{ToMe} proposed a simple token merging technique that potentially does not require retraining.

\paragraph{Robustness of ViTs.} While several Recent research has yielded mixed results on the robustness of Vision Transformers (ViTs) compared to Convolutional Neural Networks (CNNs). While some studies~\cite{understanding_robust_vit_cls,intriguing_vit,ECCVAreRobust,vit_robust} suggest ViTs are more robust against various perturbations and distribution shifts, Liu et al. \cite{ConvNeXt} challenge this notion by demonstrating that a well-designed CNN can outperform ViTs in generalization and robustness.

To enhance ViT robustness, various methods have been proposed, including network structural adjustments, patch augmentation, and diverse training strategies~\cite{RVT,FAN,RobustViT,understanding_vit_robust_aug,TORAViT,ECCVAreRobust,TAPandADL,improving_robust_vit}. For instance, Robust Vision Transformer (RVT) \cite{RVT} introduces a convolutional stem and token pooling to improve robustness, while Full Attention Net (FAN) \cite{FAN} leverages an attentional channel processing design. RobustViT \cite{RobustViT} downplays the influence of image backgrounds, and a method proposed in \cite{ECCVAreRobust} uses temperature scaling to smooth attention weights.

Additionally, Qin \textit{et al.}\cite{understanding_vit_robust_aug} improve the robustness of ViTs by using images transformed with patch-based operations as negative augmentation. Li \textit{et al.}~\cite{TORAViT} propose TORA-ViT, which consists of an accuracy adapter, a robustness adapter, and a gated fusion module. The accuracy adapter extracts predictive features, while the robustness adapter extracts robust features. These features are then combined by the gated fusion module. Reducing Sensitivity to Patch Corruptions (RSPC)~\cite{improving_robust_vit} enhances the robustness of ViTs through a specialized training strategy. In~\cite{TAPandADL}, the Attention Diversification Loss (ADL) is introduced to encourage output tokens to aggregate information from a diverse set of input tokens. However, most of these approaches require extensive training or fine-tuning and often sacrifice performance for efficiency~\cite{ToMe}. In contrast, our method can be applied to baseline vision transformers\cite{vit, DeiT} without requiring additional training and without any performance drop.

\section{Preliminaries}

\begin{figure}[t]
    \centering
    \begin{subfigure}[b]{0.28\textwidth}
        \centering
    \includegraphics[width=\textwidth]{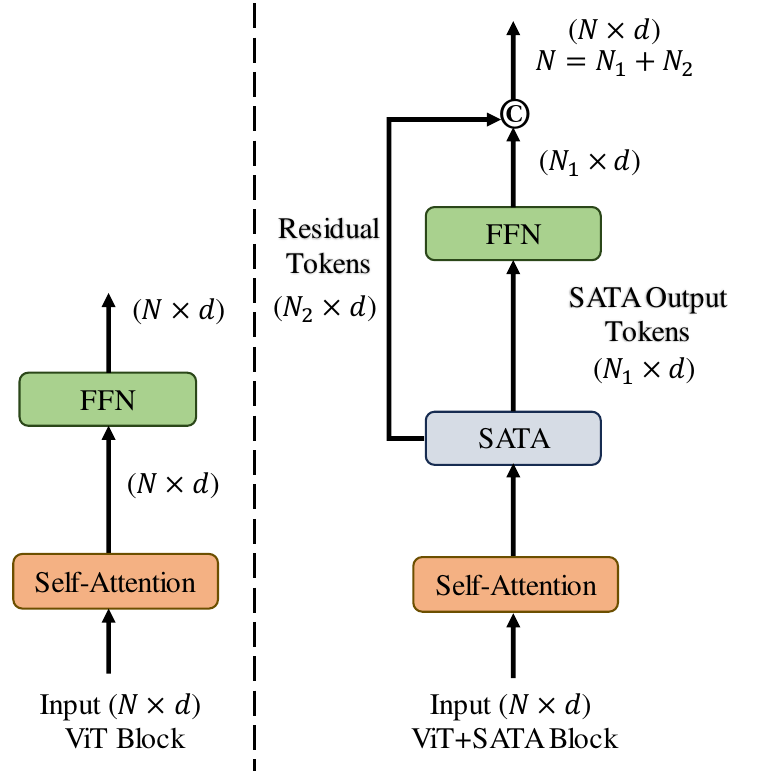}%
        \caption{}
        \label{fig:sub1}
    \end{subfigure}
    \hfill 
    \begin{subfigure}[b]{0.7\textwidth}
        \centering
        \includegraphics[width=\textwidth]{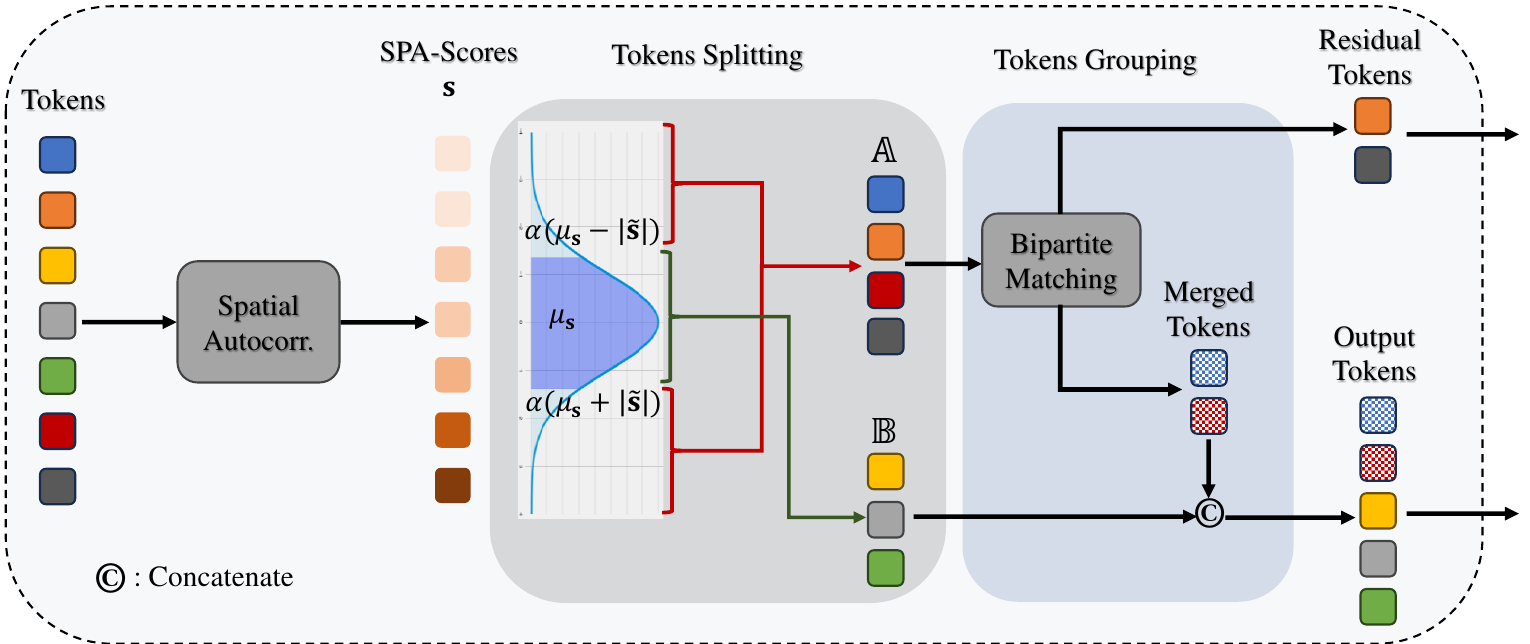}
        \caption{}
        \label{fig:sub2}
    \end{subfigure}
    \hfill
    \begin{subfigure}[b]{0.55\textwidth}
        \centering
        \includegraphics[width=\textwidth]{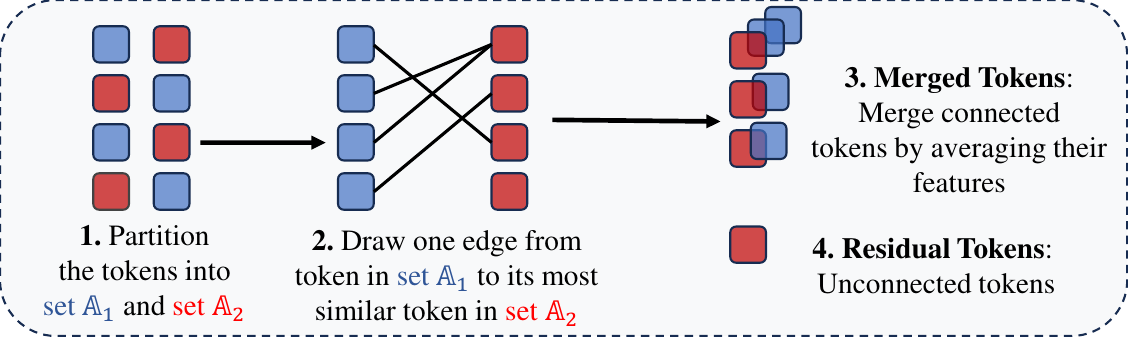}
        \caption{}
        \label{fig:sub3}
    \end{subfigure}    
    \caption{\textbf{(a)} Comparison between conventional ViT block and the augmented ViT with SATA \textbf{(b)} Overall architecture of the proposed SATA module. \textbf{(c)} Bipartite Matching.}
    \label{fig:overall-sata}
    \vspace*{-3mm}
\end{figure}

\subsection{Vision Transformers: Multi-head Self Attention (MHSA)}
A standard ViT~\cite{vit} partitions an input image into $N$ patches (tokens). These patches are then transformed to generate a token embedding tensor $\mathbf{X} \in \mathbb{R}^{N\times d}$. These tokens are then processed through a stack of transformer blocks, as illustrated in Figure 2(a). ViTs leverage self-attention~\cite{AttAll} to aggregate global information. Given the input token embedding tensor $\mathbf{X} \in \mathbb{R}^{N \times d}$, self-attention applies linear transformations with parameters $W_K$, $W_Q$, and $W_V$ to embed them into the key $K=W_K\mathbf{X}$, query $Q=W_Q\mathbf{X}$, and value $V=W_V\mathbf{X}$, respectively. Self-attention utilises $K$ and $Q$ to generate a pairwise attention map $\mathbf{M}_{att}\in \mathbb{R}^{N \times N}$ and then aggregates the token features using the attention map $\mathbf{M}_{att}$ as follows:

\begin{align}
\mathbf{M}_{att}=\text{Softmax}(QK^t/\sqrt{d}),\\
\text{Self-Attention}(Q, K, V)= \mathbf{M}_{att}V,
\label{eq:att_map}
\end{align}
where the symbol ``$t$'' indicates the transpose of the matrix. To achieve rich feature hierarchies, the Transformer block employs multiple self-attention heads. Specifically, $h$ heads are stacked in parallel, resulting in an output of $N\times h\times d$. These concatenated features are then processed by a feed-forward network (FFN) for further transformation. Finally, the FFN output of $N\times d$ serves as the final output of the Multi-Head Self-Attention (MHSA) block within the Transformer architecture.

\subsection{Geographical Spatial Auto-correlation}
In geographical modelling, spatial autocorrelation plays a crucial role in assessing the spatial interdependence of entities based on their locations and values. Positive spatial autocorrelation indicates that neighbouring observations share similar values, while negative spatial autocorrelation suggests that nearby observations tend to have contrasting values. Typically, two types of measures are used: global measures, which provide an overall assessment of spatial autocorrelation across all data points, and local measures, which offer insights into the spatial autocorrelation of individual locations relative to their neighbourhoods. Moran's metric~\cite{Moran,new_moransI} is commonly employed in geographical analysis to compute such measurements. In this study, we employ Moran's measurement for the first time, to the best of our knowledge, to investigate spatial dependency among vision transformers' tokens (patches). 

Let $\mathbf{X}$ be a set of $N$ observations (here, tokens) presented by embedded vectors $\mathbf{x}_i \in \mathbb{R}^d$, $\mathbf{X}=[\mathbf{x}_1,\mathbf{x}_2,...,\mathbf{x}_N]$, and an associated attribute, $\mathbf{a}=[a_1,a_2,...,a_N]$, the local Moran's I metric can be defined as:

\begin{equation}
 \bm{\mathit{I}}_{l}=[\text{diag}(\mathbf{z}\mathbf{z}^{t}\mathbf{W})]_{N\times 1},
\label{eq:local_new_moran}
\end{equation}
where $\text{diag}(.)$ returns the diagonal elements of a matrix. Symbol ``$t$'' indicates the transpose of the matrix. $\mathbf{W}=[w_{ij}]_{N\times N}$ represents spatial weight matrix, in which $w_{ij}$ denotes the degree of closeness or the contiguous relationships between $\mathbf{x}_i$ and $\mathbf{x}_j$ and can be computed using a dot product similarity ($\mathbf{x}_i . \mathbf{x}_{j}^t 
\text{   }i,j=1,2,...,N$). $\mathbf{z}$ refers to normalised token-wise attribute values as:

\begin{equation}
 \mathbf{z}=\frac{\mathbf{a}-\mu}{\sigma},
\label{eq:norm_gap}
\end{equation}
where $\mu$ and $\sigma$ denote mean and standard deviation of $\mathbf{a}$, respectively. The final local spatial autocorrelation descriptor, $\mathbf{s}$, can be defined as normalised $\bm{\mathit{I}}_{l}$~\cite{csa_net}:
\begin{equation}
 \mathbf{s}=\frac{\bm{\mathit{I}}_{l}-\mu_{\bm{\mathit{I}}_{l}}}{\sigma_{\bm{\mathit{I}}_{l}}},
\label{eq:spa}
\end{equation}
where $\mu_{\bm{\mathit{I}}_{l}}$ and $\sigma_{\bm{\mathit{I}}_{l}}$ indicate mean and standard deviation of $\bm{\mathit{I}}_{l}$, respectively. Following~\cite{csa_net}, given a token embedding tensor $\mathbf{X}=[\mathbf{x}_1,\mathbf{x}_2,...,\mathbf{x}_N] \in \mathbb{R}^{N\times d}$, its token-wise global context attribute $\mathbf{a}=[a_1,a_2,...,a_N]\in \mathbb{R}^{N\times 1}$ can be defined as:
\begin{align}
\mathbf{a}=\left[ a_{i}=\frac{1}{d}\sum_{t=1}^{d} \mathbf{x}_{i}(t)\right]_{N\times 1},
\label{eq:gavg}
\end{align}
where $d$ denotes the spatial dimension of the tokens. $\mathbf{x}_{i}(t)$ represents the $i$-th token value at position $t$. It's worth noting that more advanced strategies or application-specific criteria can be employed to derive the global contextual information descriptor (Eq.(\ref{eq:gavg})). In this context, we adopt the same approach as\cite{csa_net} for the sake of simplicity.

\begin{figure}[t]
  \centering
    \includegraphics[trim=0.cm 0.0cm 0.cm 0.cm, clip, width=0.99\linewidth]{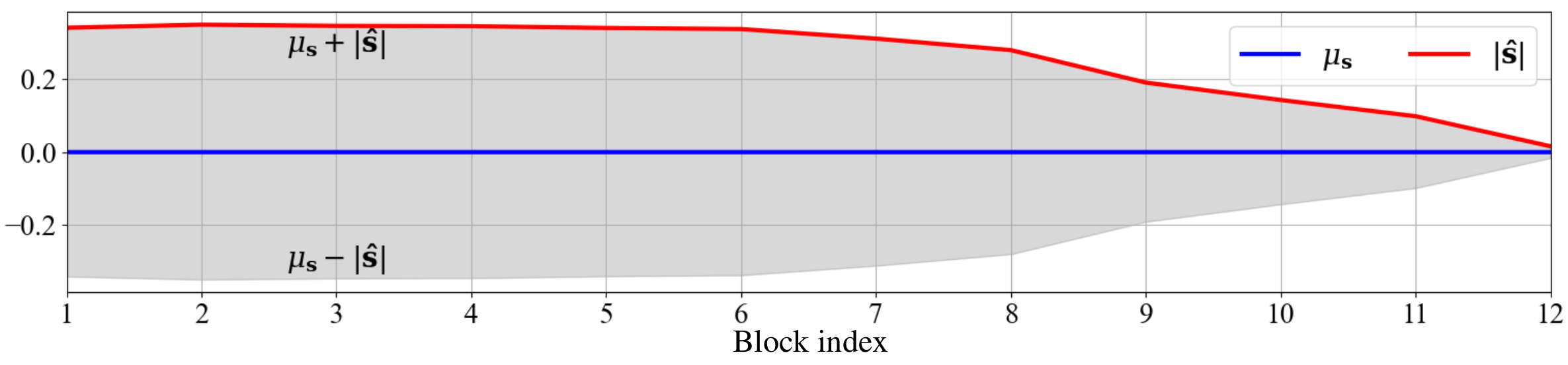}    
\caption{ Plotting the variations of $\mu_{\mathbf{s}}$, $|\mathbf{\hat{s}}|$, and the lower and upper bounds across different blocks of the ViT.}
\label{fig:spt_stage}
\vspace*{-5mm}
\end{figure}

\section{Spatial Autocorrelation Token Analysis}\label{sec:SATA}
Figure~\ref{fig:overall-sata}(b) illustrates the overall architecture of the proposed spatial autocorrelation token analysis module, situated between attention and FFN units within a standard ViT block, as depicted in Figure~\ref{fig:overall-sata}(a). To initiate our spatial autocorrelation token analysis, we begin with the observation of the spatial autocorrelation scores, $\mathbf{s}$ (Eq. (\ref{eq:spa})), for ViT's token embedding tensors over different blocks. As transformers inherently capture pairwise closeness relationships between tokens by computing the attention map $\mathbf{M}_{att}$ in Eq.(\ref{eq:att_map}), we can directly set $\mathbf{W} = \mathbf{M}_{att}$ in Eq.(\ref{eq:local_new_moran}) to enhance both efficiency and effectiveness\footnote{The implementation of the proposed SATA is provided in Appendix~\ref{subsec-apx:code}}.

 Figure~\ref{fig:spt_stage} illustrates the alterations in the mean ($\mu_{\mathbf{s}}$) and the absolute value of the median ($|\mathbf{\hat{s}}|$) statistics of the spatial autocorrelation ($\mathbf{s}$) across different blocks of Deit-Base/16 on the ImageNet-1K validation set. It can be seen that for the later layers, specifically starting from block six, tokens tend to exhibit lower $|\mathbf{\hat{s}}|$ values. Drawing from the aforementioned observation, we encapsulate the proposed analysis into two sequential steps to handle these tokens to improve the ViT's robustness and performance:

\subsection{Token Splitting}\label{subsec: tk-split}
Based on the above findings, the overall of the proposed SATA method is illustrated in Figure~\ref{fig:overall-sata} (b). As shown in Figure~\ref{fig:spt_stage}, we limit token processing to the latter stages of the transformer. Specifically, tokens in layers from $\gamma \times B$ onward ($\gamma > 0$, where $B$ represents the depth, or number of blocks, of the transformer) are partitioned into two sets, $\mathbb{A}$ and $\mathbb{B}$, using the spatial autocorrelation scores $\mathbf{s}$ as follows:

\begin{align}
\mathbb{A} &=\{\mathbf{x}_i \text{  ;  } s_i < \alpha(\mu_{\mathbf{s}} - |\mathbf{\hat{s}}|) \text{ and } s_i > \alpha(\mu_{\mathbf{s}} + |\mathbf{\hat{s}}|) \}\\
\mathbb{B} &= \{\mathbf{x}_j \text{  ;  }  \alpha(\mu_{\mathbf{s}} - |\mathbf{\hat{s}}|)<= s_j <= \alpha(\mu_{\mathbf{s}} + |\mathbf{\hat{s}}|) \}
\label{eq:upper-lower}
\end{align}

where $\alpha(\mu_{\mathbf{s}} - |\mathbf{\hat{s}}|)$ and  $\alpha(\mu_{\mathbf{s}} + |\mathbf{\hat{s}}|)$ represent lower and upper bounds, respectively. $\alpha$ denotes the controlling factor (parameters $\alpha$ and $\gamma$ choices are discussed in Section~\ref{sec:abl-study}).    

\subsection{Token Grouping}\label{sec:token-group}
To manage tokens falling beyond the lower and upper bounds (\textit{i.e.} $\alpha(\mu_{\mathbf{s}} \pm |\mathbf{\hat{s}}|)$), we employ the Bipartite Matching algorithm~\cite{ToMe}, to efficiently match and merge similar tokens in set $\mathbb{A}$. In particular, the Bipartite Matching algorithm can be summarized as follows (illustrated in Figure~\ref{fig:overall-sata}(c)):
\begin{enumerate}[itemsep=0pt, parsep=0pt, topsep=0pt]
    \item Partition set $\mathbb{A}$ into two sets $\mathbb{A}_1$ and $\mathbb{A}_2$ of roughly equal size.
    \item Draw one edge from each token in $\mathbb{A}_1$ to its most similar token in $\mathbb{A}_2$.
    \item \textbf{Merged Tokens}: Merge connected tokens by averaging their features
    \item \textbf{Residual Tokens}: Unconnected tokens
\end{enumerate}
As depicted in Figure~\ref{fig:overall-sata}(b), the output tokens of the proposed SATA module comprise the concatenation of tokens with spatial scores within the range of lower and upper bounds (\textit{i.e.}, set $\mathbb{B}$) and the \textbf{Merged Tokens} resulting from the Bipartite Matching algorithm, which is then fed into the FFN module. Additionally, the \textbf{Residual Tokens} are concatenated with the output of the FFN to form the final output of the new ViT block and restore the original number of tokens, $N$.


\section{Experiment Results \& Analysis}
\definecolor{lightgray}{gray}{0.9}
\begin{table}
\caption{Performance of SATA and several state-of-the-art (SOTA) CNNs and ViTs models on ImageNet and six robustness benchmarks: We report the mean corruption error (mCE) for ImageNet-C~\cite{ImageNet-C}, where lower mCE values indicate higher model robustness. Our SATA models consistently outperform other counterparts in standard performance and enhance robustness across various model sizes compared to the baseline, all without requiring additional training or fine-tuning. SATA-B$^*$ refers to the integration of the proposed SATA module into the pre-trained vanilla ViT-Base/16 model~\cite{vit}.}
\resizebox{\columnwidth}{!}{%
\centering
\begin{tabular}{cccccccccccc}
\toprule
\multirow{2}{*}{Group}&\multirow{2}{*}{Model}&FLOPs&Params&\multicolumn{2}{c}{ImageNet-1K} & \multicolumn{6}{c}{Robustness Benchmarks}\\
&&(G)&(M)&Top-1&Top-5&FGSM&PGD&IN-C(mCE$\downarrow$)&IN-A&IN-R&IN-SK\\
\midrule

\multirow{7}{*}{CNN}
 & ResNet50~\cite{he2016deep}   & 4.1 & 25.6& 76.1& 86.0& 12.2&0.9&76.7&0.0&36.1&24.1\\
 & RegNetY-4GF\cite{RegNet} & 4.0 & 20.6& 79.2& 94.7& 15.4&2.4&68.7&8.9&38.8&25.9\\
 & EfficientNet-B4\cite{EfficientNet}   & 4.4 &19.3& 83.0& 96.3& 44.6&18.5&71.1&26.3&47.1&34.1\\
 & DeepAugment\cite{DeepAug}  & 4.1 & 25.6& 75.8& 92.7&27.1&9.5&53.6&3.9&46.7&32.6\\
 & ANT\cite{ANT}  & 4.1 & 25.6& 76.1& 93.0&17.8&3.1&63.0&1.1&39.0&26.3\\
 & Debiased CNN\cite{DebiasedCNN}  & 4.1 & 25.6& 76.9& 93.4&20.4&5.5&67.5&3.5&40.8&28.4\\
 & ConvNeXt-B\cite{ConvNeXt}  & 15.4 & 89& 83.8& -&-&-&46.8&36.7&51.3&38.2\\\hline
\multirow{8}{*}{ViT-Tiny} 
 & DeiT-Ti\cite{DeiT} & 1.3 & 5.7& 72.2& 91.1& 22.3&6.2&71.1&7.3&32.6&20.2\\
 & ConvViT-Ti\cite{Convit}& 1.4 & 5.7& 73.3& 91.8& 24.7&7.5&68.4&8.9&35.2&22.4\\ 
 & PiT-Ti\cite{PiT}    & 0.7 & 4.9& 72.9& 91.3& 20.4&5.1&69.1&6.2&34.6&21.6\\
 & PVT-Tiny\cite{PVT}  & 1.9 & 13.2& 75.0& 92.5&10.0&0.5&79.6&7.9&33.9&21.5\\
 & RVT-Ti~\cite{RVT}  & 1.3 & 8.6& 78.4& 94.2&34.8&11.7&58.2&13.3&43.7&\textbf{30.0}\\
 & FAN-T-ViT~\cite{FAN}  & 1.3 & 7.0& 79.2& -&-&-&57.5&-&42.5&-\\
 & RVT-Ti+RSPC~\cite{improving_robust_vit}  & 1.3 & 10.9& 79.2& -&-&-&55.7&\textbf{16.5}&-&-\\
 \rowcolor{lightgray}
 & SATA-T (ours)  & 1.0 & 5.7& \textbf{86.5}& \textbf{98.2}&\textbf{40.0}&10.9&\textbf{51.1}&\uline{14.6}&\textbf{47.3}&\uline{25.2}\\\hline
\multirow{11}{*}{ViT-Small}  
& DeiT-S\cite{DeiT}   &4.6 & 22.1& 79.9& 95.0& 40.7&16.7&54.6&18.9&42.2&29.4\\
& ConvViT-S\cite{Convit}&5.4 & 27.8& 81.5& 95.8& 41.7&17.2&49.8&24.5&45.4&33.1\\
& PiT-S\cite{PiT}    &2.9 & 23.5& 80.9& 95.3& 41.0&16.5&52.5&21.7&43.6&30.8\\
& PVT-Small\cite{PVT}&3.8 & 24.5& 79.9& 95.0& 26.6&3.1&66.9&18.0&40.1&27.2\\
& Swin-T\cite{SwinViT}   &4.5 & 28.3& 81.2& 95.5& 33.7&7.3&62.0&21.6&41.3&29.1\\
& TNT-S\cite{TNT}    &5.2 & 23.8& 81.5& 95.7& 33.2&4.2&53.1&24.7&43.8&31.6\\
& T2T-ViT\_t-14\cite{T2T-ViT}& 6.1 & 21.5& 81.7& 95.9& 40.9&11.7&53.2&23.9&45.0&32.5\\
 & RVT-S~\cite{RVT}  & 4.7 & 22.1& 81.7& 95.7&51.3&26.0&50.1&24.1&46.9&35.0\\
 & FAN-S-ViT~\cite{FAN}  & 5.3 & 28.0& 82.9& -&-&-&47.7&29.1&50.4&-\\
 & RVT-S+RSPC~\cite{improving_robust_vit}  & 4.7 & 23.3& 82.2& -&-&-&48.4&27.9&-&-\\
\rowcolor{lightgray}
& SATA-S (ours)  & 3.9 & 22.1& \textbf{89.3}& \textbf{99.1}&\textbf{57.4}&18.0&\textbf{33.8} &\textbf{30.5}&\textbf{59.5}&\textbf{39.2}\\\hline
\multirow{13}{*}{ViT-Base}  
& DeiT-B\cite{DeiT}   &17.6&86.6&82.0&95.7&46.4&21.3&48.5&27.4&44.9&32.4\\
& ConvViT-B\cite{Convit}&17.7&86.5&82.0&95.7&46.4&21.3&48.5&27.4&44.9&32.4\\
& PiT-B\cite{PiT}    &12.5&73.8&82.4&95.7& 49.3&23.7&48.2&33.9&43.7&32.3\\
& PVT-Large\cite{PVT}&9.8 &61.4&81.7&95.9& 33.1&7.3&59.8&26.6&42.7&30.2\\
& Swin-B\cite{SwinViT}   &15.4&87.8&83.4&96.4& 49.2&21.3&54.4&35.8&46.6&32.4\\
& T2T-ViT\_t-24\cite{T2T-ViT}& 15.0&64.1&82.6&96.1&46.7&17.5&48.4&28.9&47.9&35.4\\
 & RVT-B~\cite{RVT}  & 17.7 & 86.2& 82.5& 96.0&52.3&27.4&47.3&27.7&48.2&35.8\\
 & FAN-B-ViT~\cite{FAN}  & 10.4& 54.0& 83.6& -&-&-&44.4&35.4&51.8&-\\
 & RVT-B+RSPC~\cite{improving_robust_vit}  & 17.7 & 91.8& 82.8& -&-&-&45.7&32.1&-&-\\
 & TORA-ViT-B/16($\lambda=0.1$)~\cite{TORAViT}  & 26.0 & 111.2& 84.1& -&48.4&23.3&31.7&46.5&57.6&-\\
 & RVT-B+RSPC~\cite{improving_robust_vit}  & 17.7 & 91.8& 82.8& -&-&-&45.7&32.1&-&-\\
\rowcolor{lightgray}
& SATA-B (ours)  & 15.9 & 86.6& \textbf{93.9}& \textbf{99.7}&\textbf{63.9}&20.2&\textbf{28.7}&\textbf{63.5}&\textbf{70.0}&\textbf{49.8}\\
\rowcolor{lightgray}
& SATA-B$^*$ (ours)  & 15.9 & 86.6& \textbf{94.9}& \textbf{99.8}&\textbf{65.6}&\textbf{28.3}&\textbf{13.6}&\textbf{63.6}&\textbf{79.2}&\textbf{57.9}
\\\bottomrule
\end{tabular}}
 \label{tbl:results}
\vspace*{-4mm}
\end{table}
\subsection{Experimental setup}\label{subsec:exp-setup}
\paragraph{Implementation Details} All experiments were conducted on an NVIDIA V100 GPU with a $224 \times 224$ image resolution. We integrated the proposed SATA module into pre-trained generic vision transformers~\cite{DeiT,vit} (Deit-Tiny/16, Deit-Small/16, Deit-Base/16, and Vit-Base/16), resulting in three model sizes named SATA-T, SATA-S, SATA-B, and SATA-B$^*$, respectively.
\paragraph{Evaluation Benchmarks} We employ the ImageNet-1K~\cite{ImageNet} dataset for standard performance evaluation. For robustness assessment, we evaluate the proposed SATA in three dimensions: $1)$ Adversarial Robustness: Testing is conducted on adversarial examples generated by white-box attack algorithms FGSM~\cite{FGSM} and PGD~\cite{PGD} using the ImageNet-1K validation set. ImageNet-A~\cite{ImageNet-A} (IN-A) includes the ImageNet objects in unusual contexts or orientations and is utilized to assess model performance against natural adversarial examples. $2)$ Common Corruption Robustness: We use ImageNet-C~\cite{ImageNet-C} (IN-C), which includes 15 types of algorithmically generated corruptions, each with five levels of severity. $3)$ Out-of-Distribution Robustness: Evaluation is performed on ImageNet-R~\cite{ImageNet-R} (IN-R) and ImageNet-Sketch~\cite{ImageNet-SK} (IN-SK). Both datasets feature images with naturally occurring distribution shifts. ImageNet-R~\cite{ImageNet-R} (IN-R) contains abstract or rendered versions of the objects. ImageNet-Sketch~\cite{ImageNet-SK} consists solely of sketch images, serving to test classification capability when texture or colour information is absent.
\subsection{Standard Performance Evaluation}
For standard performance evaluation, we compare our method with several state-of-the-art classification models, including Transformer-based models and representative CNN-based models, as shown in Table~\ref{tbl:results}. Our proposed SATA significantly outperforms all other architectures, including both CNN-based and ViT-based models. Specifically, ViT models enhanced with SATA achieve new state-of-the-art top-1 accuracy of 86.5\%, 89.3\%, and 93.9\% for the tiny, small, and base versions, respectively, all without requiring additional training, input augmentation, or fine-tuning.  Notably, integrating the proposed SATA into pre-trained ViT-Base/16~\cite{vit} (SATA-B$^*$) results in an additional 1.0\% improvement. Furthermore, comparing the computation cost (GFLOPs) of the baseline DeiTs and SATA models demonstrates that the proposed spatial autocorrelation token analysis method also improves efficiency.
\subsection{Adversarial Robustness Evaluation}
For evaluating white-box attack adversarial robustness, we follow~\cite{RVT} and adopt the single-step attack algorithm FGSM~\cite{FGSM} and the multi-step attack algorithm PGD~\cite{PGD} (with 5 steps and a step size of 0.5). Both attackers perturb the input image with a maximum magnitude of $\epsilon = 1$. As shown in Table~\ref{tbl:results}, adversarial robustness appears unrelated to standard performance. For instance, models like Swin~\cite{SwinViT}, PVT~\cite{PVT}, and TNT-S~\cite{TNT} achieve higher standard accuracy than DeiTs corresponding, but their adversarial robustness is significantly lower, consistent with findings from~\cite{RVT, on_adversarial_rb_vit}. Our proposed SATA model achieves superior performance against both FGSM~\cite{FGSM} and PGD~\cite{PGD} attacks. Specifically, SATA-S, SATA-B, and SATA-B$^*$ show over a 20\% improvement on FGSM~\cite{FGSM} compared to previous ViT variants.

Regarding natural adversarial robustness, the proposed SATA-T demonstrates a comparable performance of 14.6\%, which is on par with some current state-of-the-art methods like RVT-Ti~\cite{RVT} and RVT-Ti+RSPC~\cite{improving_robust_vit}, while being about half their size. However, for models of similar size (e.g., ViT-Small and ViT-Base), the proposed SATAs outperform others by about 50\%, indicating the effectiveness of SATA against natural adversarial attacks.
\subsection{Common Corruption Robustness Evaluation}
To measure model degradation on common image corruptions, we report the mean corruption error (mCE) on ImageNet-C~\cite{ImageNet-C} (IN-C) in Table~\ref{tbl:results}. Our SATA method significantly reduces the mCE of DeiT-Ti~\cite{DeiT} from 71.1\% to 51.1\%, achieving the lowest mCE among vision transformers within the ViT-Tiny group. For the other two larger ViT groups, the proposed SATA models achieve an mCE of approximately 28\%, improving by around 20 points over all other ViT or CNN-based methods on the leaderboard, thereby establishing a new state-of-the-art. This result also suggests that spatial autocorrelation management of visual tokens can successfully handle different types of image corruption.
\subsection{Out-of-Distribution Robustness Evaluation}
We evaluate the generalization ability of SATA on out-of-distribution data by reporting the top-1 accuracy on ImageNet-R~\cite{ImageNet-R} (IN-R) and ImageNet-Sketch~\cite{ImageNet-SK} (IN-SK) in Table~\ref{tbl:results}. The generic vision transformers~\cite{vit,DeiT} enhanced by the proposed SATA consistently outperform other ViT models on ImageNet-R~\cite{ImageNet-R}, achieving 47.3\%, 57.2\%, 70.0\%, 79.9\% in the ViT-Tiny, ViT-Small, and ViT-Base groups, respectively. Regarding ImageNet-Sketch~\cite{ImageNet-SK} (IN-SK), SATA demonstrates superior performance compared to other models of similar size. These results imply that the spatial autocorrelation tokens analysis effectively captures feature distribution shifts, enhancing the model's out-of-distribution generalization capabilities.
\subsection{Ablation study}\label{sec:abl-study}
\begin{figure}
\centering   
  \begin{subfigure}{0.47\textwidth}
    \centering
    \includegraphics[trim=2cm 0.cm 2.cm 0.5cm, clip, width=\textwidth ]{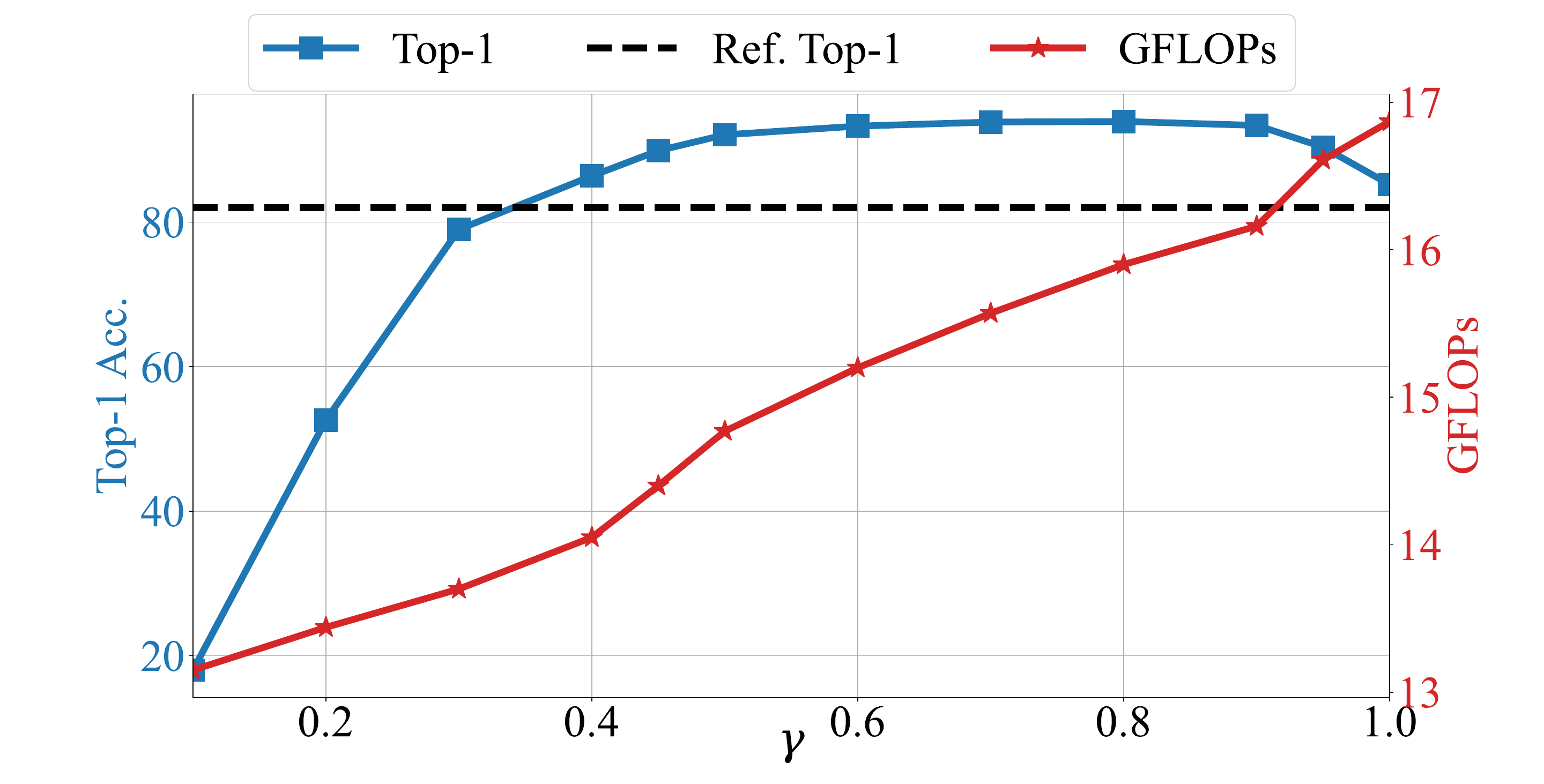}
    \caption{}
  \end{subfigure}
  \hfill
  \begin{subfigure}{0.47\textwidth}
    \centering
    \includegraphics[trim=1cm 0.cm 1.cm 0.5cm, clip, width=\textwidth ]{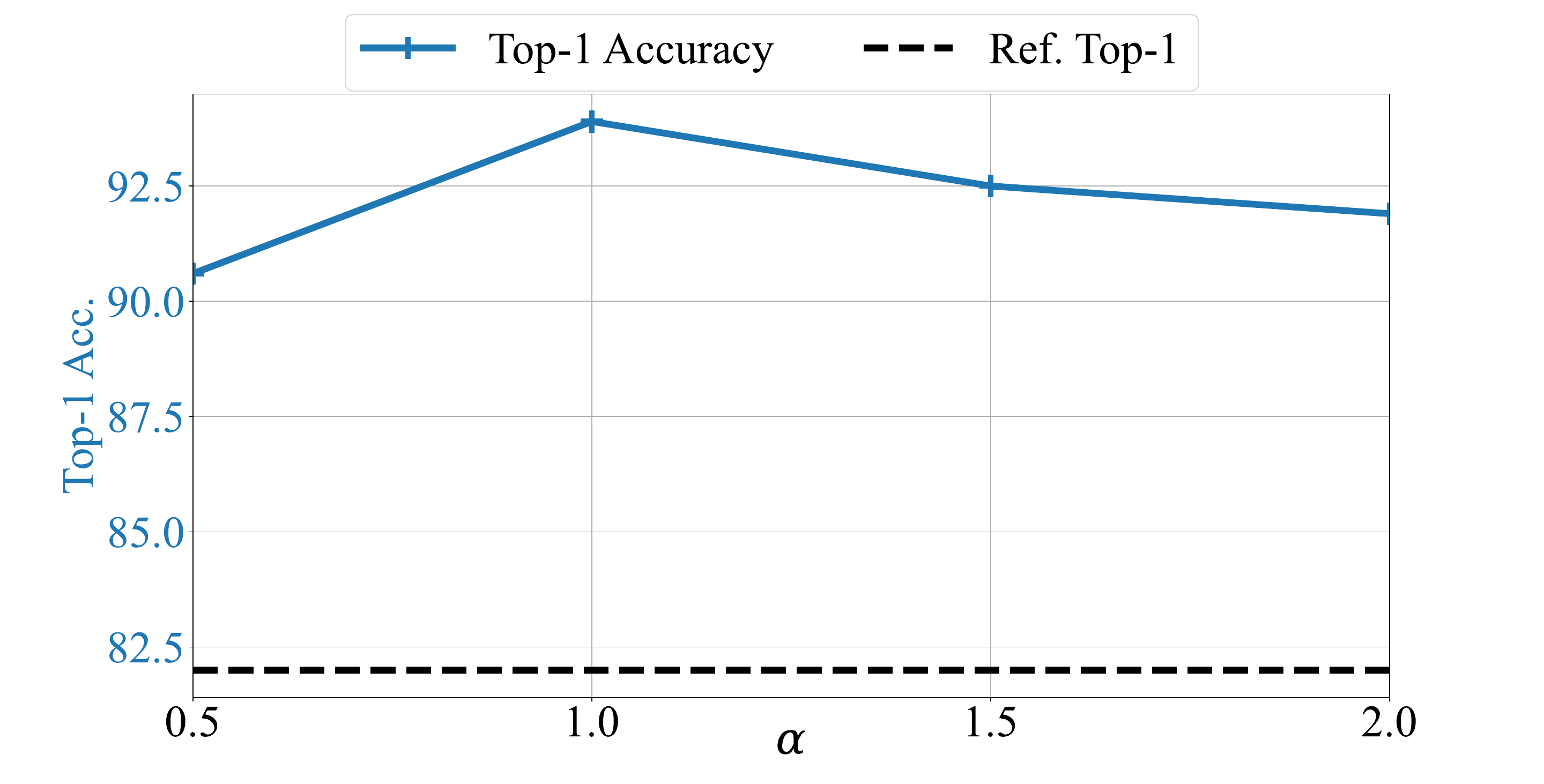}
    \caption{}
  \end{subfigure}
  \caption{\textbf{(a)} Ablation study on $\gamma$. \textbf{(b)} Ablation study on $\alpha$. We set $\gamma$ and $\alpha$ to 0.7 and 1.0, respectively, for all experiments throughout this paper. Ablation studies are conducted on the SATA-B model using the ImageNet-1K dataset. Dashed lines for both graphs represent the baseline Deit-Base/16 top-1 accuracy.}
  \label{fig:gamma_alpha}
  \vspace*{-4mm}
\end{figure}
\begin{table}[t]
\caption{Ablation study of the spatial autocorrelation module (token splitting), and token grouping (bipartite matching). The symbols ``\cmark'' and ``\xmark'' indicate whether the corresponding element is employed with the configuration or not. $\dagger$ represents top-1 accuracy for the baseline Deit-Base/16~\cite{DeiT} model.}
\centering
\begin{tabular}{cccc}
\toprule
\multicolumn{2}{c}{Spt. Auto. Correlation (Tokens Splitting)} & Tokens Grouping & \multirow{2}{*}{Top-1}\\
Lower bound ($\mu_{\mathbf{s}} - |\mathbf{\hat{s}}|$)& Upper bound ($\mu_{\mathbf{s}} + |\mathbf{\hat{s}}|$)& Bipartite Matching&\\
\midrule
\xmark & \xmark& \xmark& 82.0$^{\dagger}$\\
 \xmark & \xmark   & \cmark & 84.4\\
 \xmark& \cmark   & \cmark & 92.8\\
 \cmark& \xmark   &\cmark  & 92.5\\
  \cmark&\cmark   & \xmark & 92.3\\
\cmark  &\cmark    &\cmark  & \textbf{93.9}\\
\bottomrule
\end{tabular}
 \label{tbl:split-group}
\vspace*{-2mm}
\end{table}
\paragraph{Token Splitting and Token Grouping}
We evaluate the role of token splitting based on the upper and lower bounds of spatial autocorrelation scores and token grouping (bipartite merging) modules. To this end, we examine five SATA configurations as depicted in Table~\ref{tbl:split-group}. As shown in Table~\ref{tbl:split-group}, utilizing only token grouping yields a top-1 accuracy of 84.4\%, which still improves over the reference (DeiT-Base) accuracy by 2.4\%. Including either lower or upper bounds significantly improves accuracy by about 10\% of the baseline top-1 accuracy, highlighting the effectiveness of the proposed spatial autocorrelation token splitting schema. The upper bound contributes slightly more, suggesting that tokens with extremely high spatial autocorrelation scores are more likely to be filtered by the splitting process. Finally, adding token grouping yields a further 1.6\% improvement over the splitting process alone.
\paragraph{Threshold of starting block ($\gamma$)}
We also examine the effect of parameter $\gamma$, which controls at which transformer block the SATA module is applied. As Figure~\ref{fig:gamma_alpha}(a) shows, applying SATA from block$0.4\times B$ onwards significantly improves model efficiency while exceeding baseline ViT accuracy (82\%). Applying SATA to earlier blocks ($\gamma < 0.4$) degrades accuracy, suggesting that high spatial correlation within token features and the importance of all tokens in early layers is beneficial. To achieve a good trade-off between accuracy and efficiency, we use $\gamma = 0.7$ in all our experiments. This results in a top-1 accuracy of 93.9\% and GFLOPs of 15.9.
\paragraph{Lower/Upper bounds controlling factor ($\alpha$)}
We further assess the influence of $\alpha$, the factor controlling the lower and upper bounds in SATA. Figure \ref{fig:gamma_alpha}(b) shows the performance of SATA-B on ImageNet-1K with $\alpha$ values ranging from 0.5 to 2. $\alpha$ determines the number of tokens passed to the FFN block and setting $\alpha=1$ yields optimal performance.
\begin{figure}
\centering   
  \begin{subfigure}[c]{0.47\textwidth}
    \centering
    \includegraphics[trim=2.5cm 0.5cm 2.cm 0.1cm, clip, width=\textwidth ]{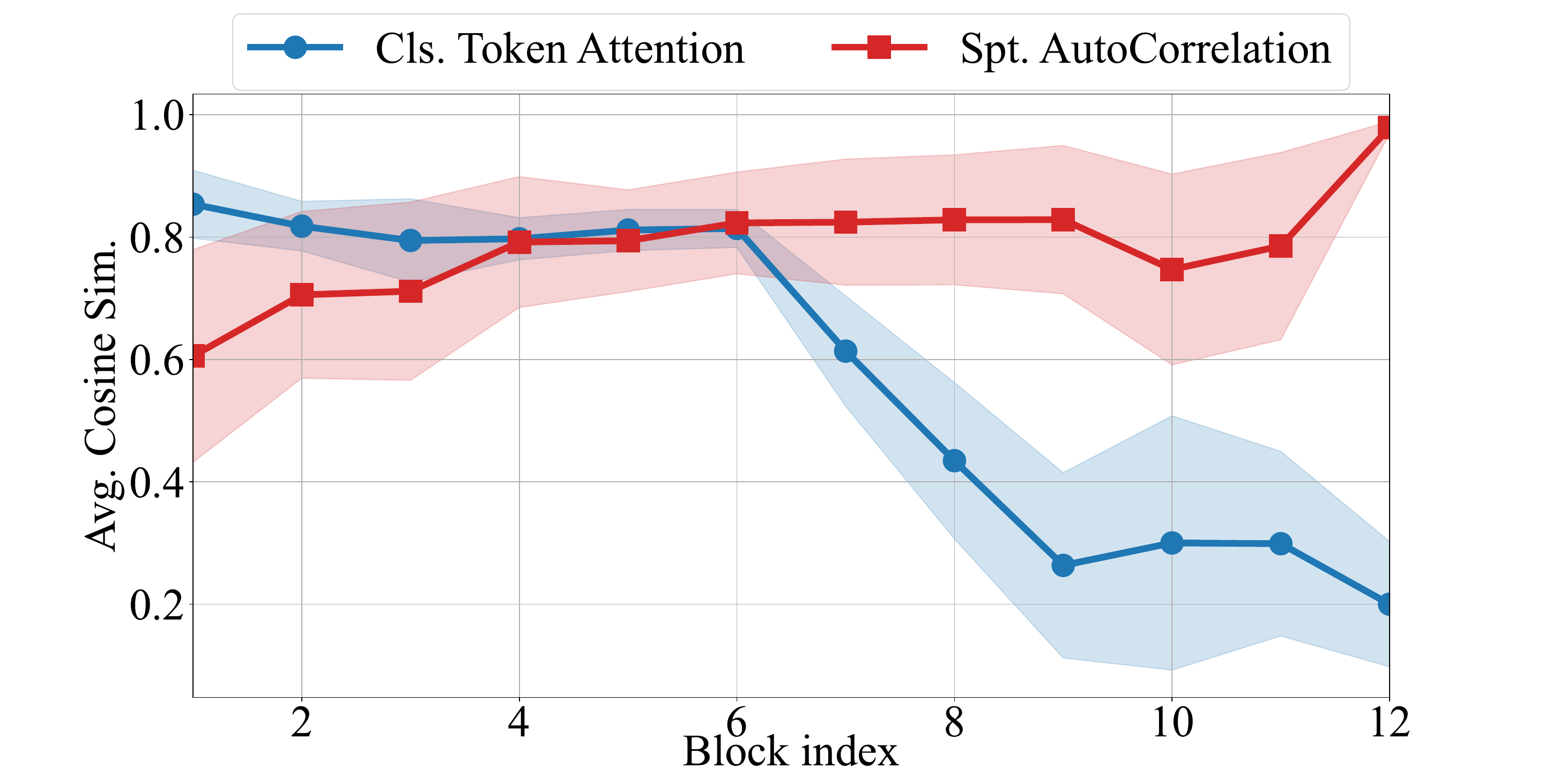}
    \caption{}
  \end{subfigure}
  \hfill
  \begin{subfigure}[c]{0.47\textwidth}
    \centering
    \includegraphics[trim=2.10cm 1.cm 1.80cm 0.10cm, clip, width=\textwidth ]{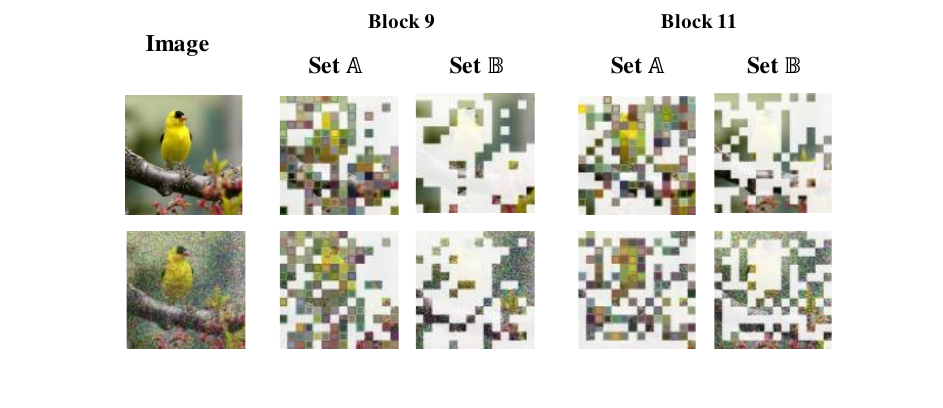}
    \caption{}
  \end{subfigure}
  \caption{\textbf{(a)} Cosine similarity between the clean and corrupted versions of the class token attention map and spatial autocorrelation scores across different blocks of SATA-B. Results are averaged across various types of image corruptions and severity levels on ImageNet-C~\cite{ImageNet-C}. \textbf{(b)} Visualisation of token splitting for a pair of clean and noisy images. Notably, the selected tokens for each set are similar for both clean and corrupted inputs.}
  \label{fig:attn_spa_AB}
    \vspace*{-4mm}
\end{figure}
\subsection{Visualisation and Discussion}
Although the effectiveness of the proposed SATA module has been empirically demonstrated, we conduct a deeper investigation to better understand its behaviour. To this end, we calculate the cosine similarity between the class token attention maps and spatial autocorrelation scores of clean and corrupted image pairs from the ImageNet-1K validation set and its corresponding ImageNet-C~\cite{ImageNet-C}, respectively. We compute these similarities for various types of image corruptions and severity levels in ImageNet-C~\cite{ImageNet-C}, and report the average values across different blocks of the proposed SATA-B in Figure~\ref{fig:attn_spa_AB}(a). As shown in Figure~\ref{fig:attn_spa_AB}(a), the cosine similarity between clean and corrupted attention maps drops significantly in the later blocks of the transformer. In contrast, the similarity for spatial autocorrelation scores improves at the early stages and remains consistently high, averaging above 0.8. This highlights that the proposed method can provide a more stable and reliable feature representation throughout the network, offering strong robustness against various types of corruption.

Moreover, Figure~\ref{fig:attn_spa_AB} provides a visualization of tokens (patches) are split into set $\mathbb{A}$ and set $\mathbb{B}$ for a pair of clean and noisy images according to the proposed SATA algorithm. Notably, the similarity between corresponding sets for clean and noisy inputs is evident, further highlighting the robustness of the proposed method\footnote{Additional enlarged visual comparisons are included in Appendix Figures~\ref{fig:setAB_apx} and \ref{fig:attn_spa_apx}.}.
\subsection{Conclusion}
In this paper, we introduce SATA, a novel method designed to significantly enhance the performance and robustness of vision transformers against various types of corruption. SATA employs a straightforward yet powerful spatial autocorrelation scheme to exploit spatial inter-dependencies among token features, thereby substantially improving representational capacity, robustness, and efficiency in terms of reducing computational costs. Our experimental results show that SATA-enhanced vision transformers consistently deliver stable and reliable feature representations, achieving state-of-the-art performance on ImageNet-1K classification and setting new benchmarks for robustness across multiple evaluations, all without the need for additional training or fine-tuning.

This work underscores SATA's transformative potential and opens several promising avenues for future research and development. Key directions include adapting SATA for window-based and hybrid ViT architectures to boost performance in tasks such as object detection and segmentation. Furthermore, exploring the application of SATA in other transformer-based domains, such as large language models (LLMs), could extend its impact even further.

\bibliography{references.bib}
\bibliographystyle{abbrvnat}
\newpage
\appendix
\section{Appendix / supplemental material}

\subsection{ImageNet-1K SOTA}
\begin{figure}[h]
    \centering
    \includegraphics[trim=4cm 0.0cm 3cm 2cm, clip, width=\linewidth]{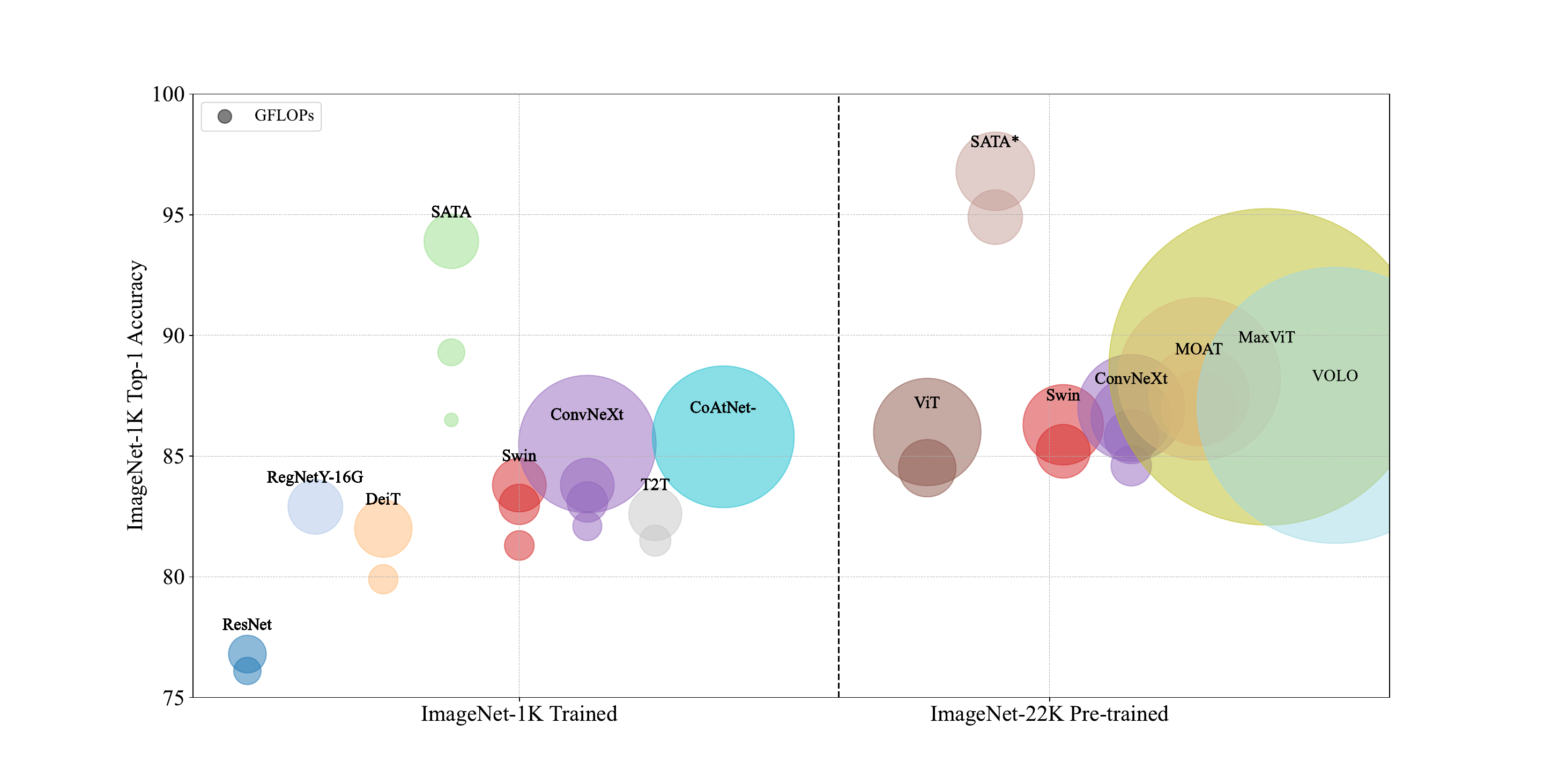}
    \caption{ImageNet-1K classification results for Vision Transformers and ConvNets models. Each bubble's area corresponds to the computational cost (GFLOPs) of a variant within its model family. ImageNet-1K/22K models use $224 \times 224$ input image resolutions, respectively. Notably, our proposed SATA and SATA$^*$ models significantly enhance the performance of standard DeiT and ViT models, establishing a new state-of-the-art.}
    \label{fig:imagenet-sota}
\end{figure}
We compare the proposed SATA models against state-of-the-art image classification models. In addition to the methods discussed in the main text, we include several current state-of-the-art models such as VOLO~\cite{VOLO}, MOAT~\cite{moat}, CoAtNet~\cite{CoAtNet}, and MaxViT~\cite{maxvit}. We compare their top-1 accuracy and efficiency in terms of GFLOPs. Notably, as shown in Figure~\ref{fig:imagenet-sota}, our proposed SATA and SATA$^*$ models significantly enhance the performance of standard DeiT and ViT models, establishing a new state-of-the-art.

\subsection{Spatial Autocorrelation distribution across ViT's blocks}
\begin{figure}[t]
  \centering
  \begin{subfigure}{\textwidth}
    \centering
    \includegraphics[trim=2.75cm 0.0cm 1.75cm 1.7cm, clip, width=0.245\linewidth]{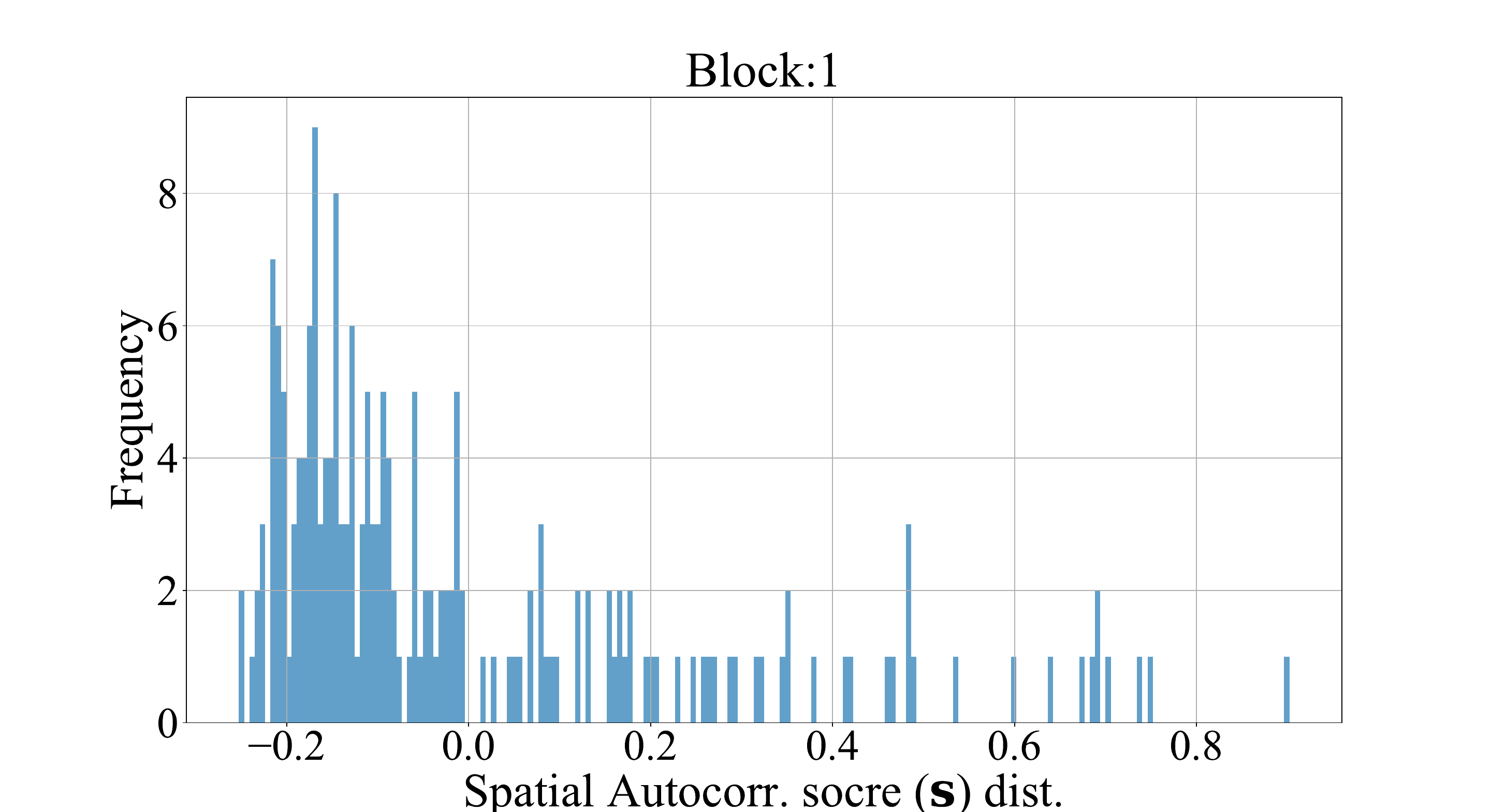}
      \hfill
    \centering
    \includegraphics[trim=2.75cm 0.0cm 1.75cm 1.7cm, clip, width=0.245\linewidth]{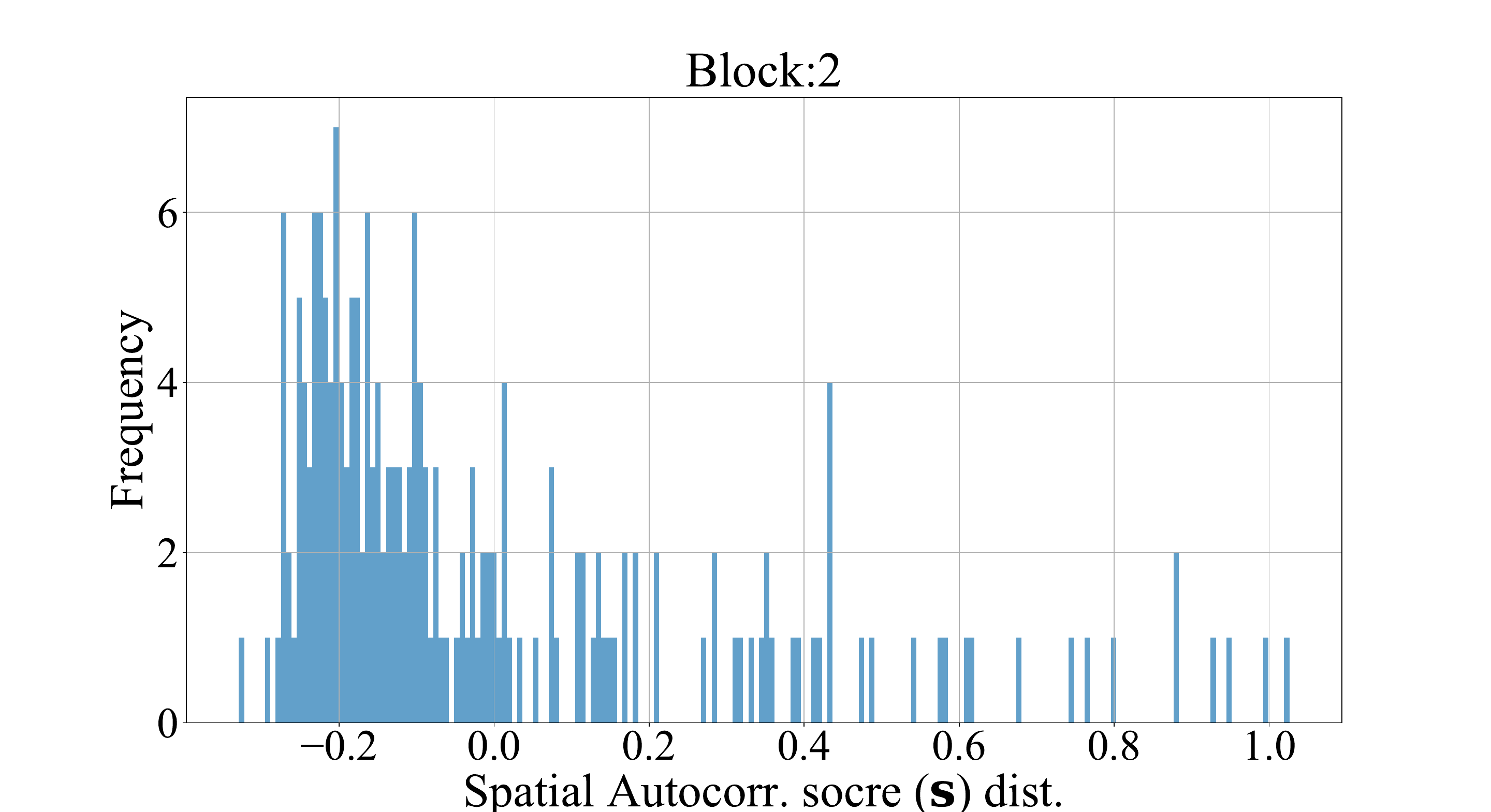}
      \hfill
    \centering
  \includegraphics[trim=2.75cm 0.0cm 1.75cm 1.7cm, clip, width=0.245\linewidth]{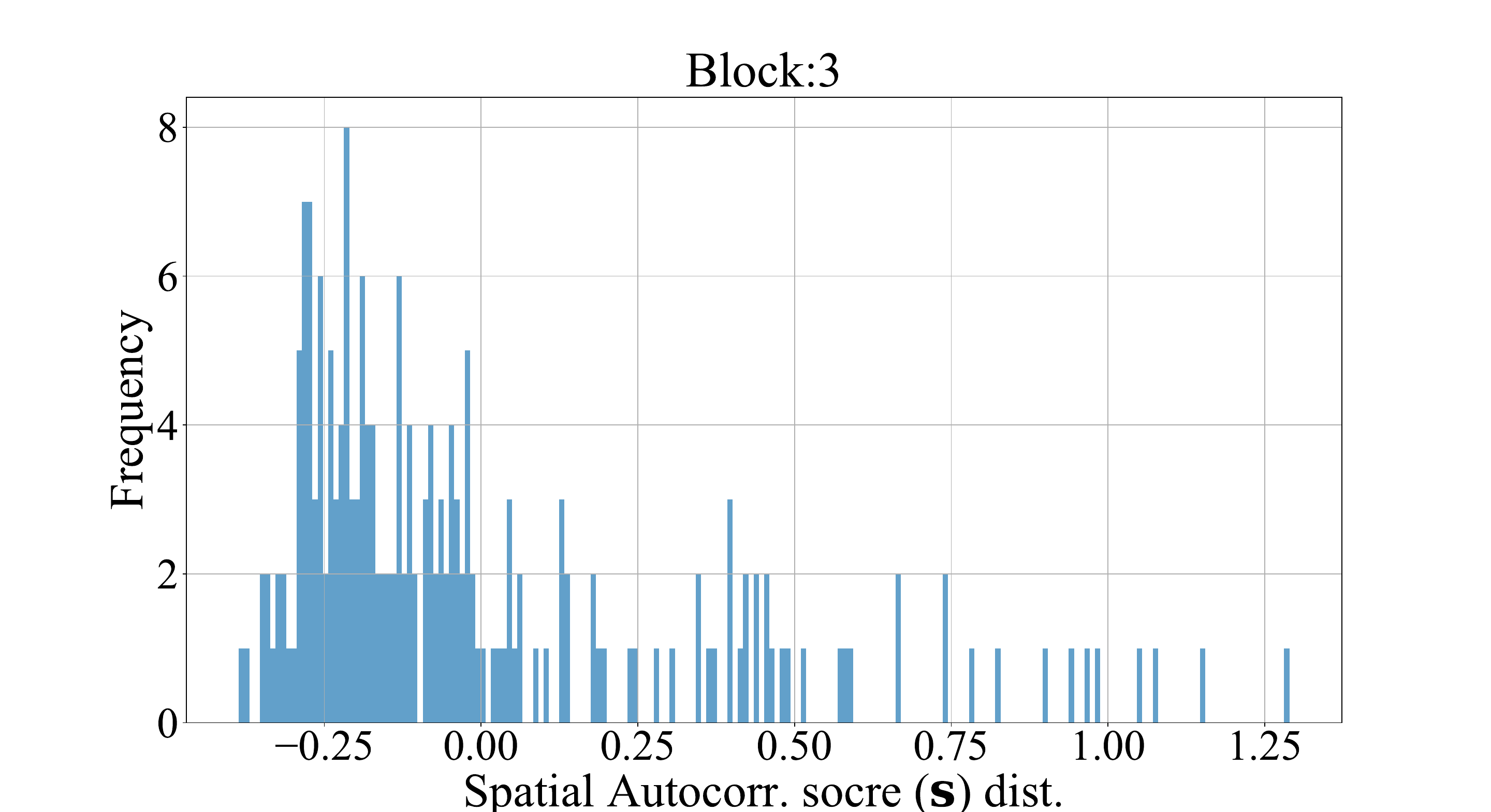}
    \hfill
    \centering
  \includegraphics[trim=2.75cm 0.0cm 1.75cm 1.7cm, clip, width=0.245\linewidth]{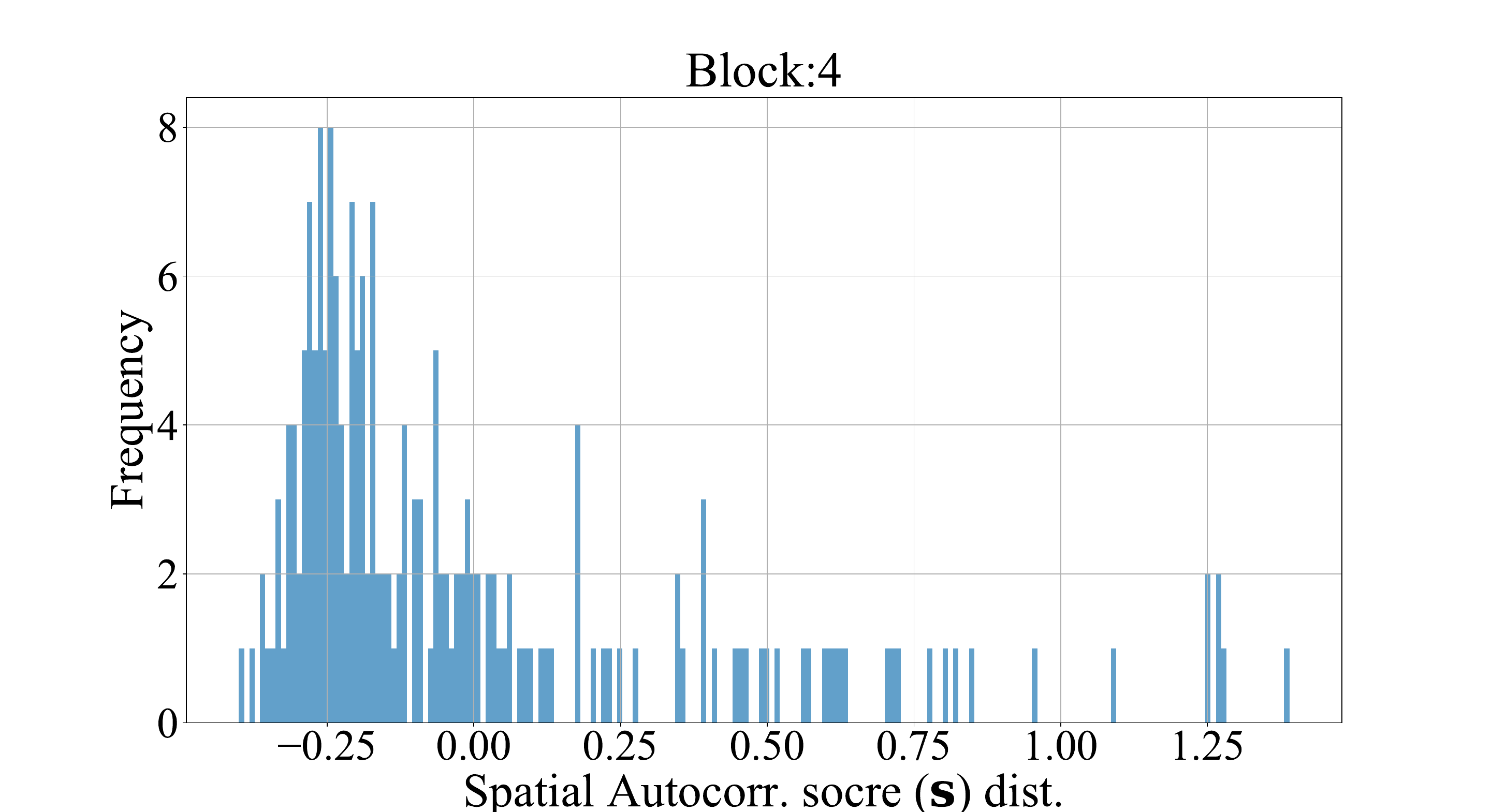}  
  \end{subfigure}

  \vspace{0.5em} 

  \begin{subfigure}{\textwidth}
    \centering
    \includegraphics[trim=2.75cm 0.0cm 1.75cm 1.7cm, clip, width=0.245\linewidth]{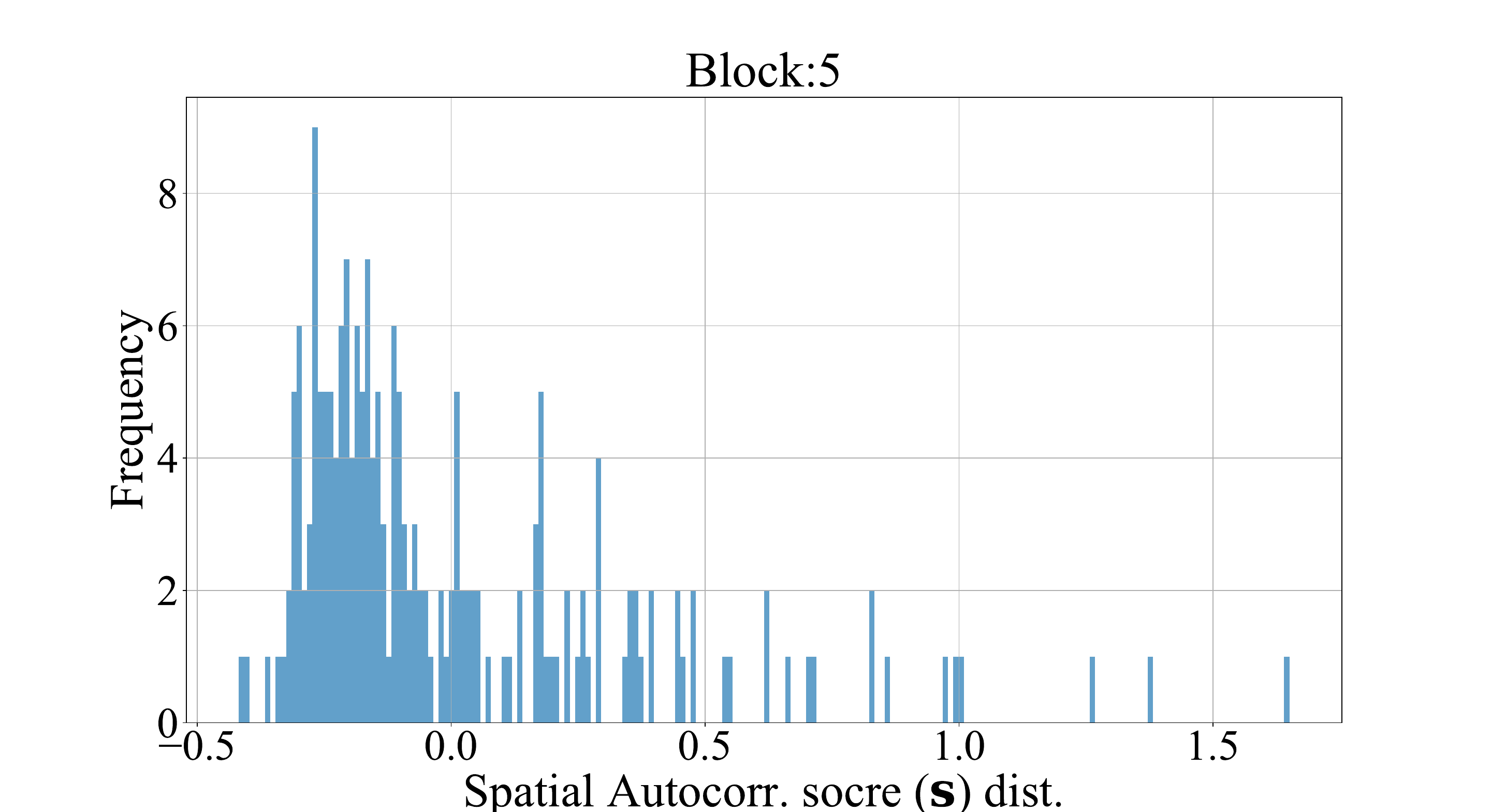}    
  \hfill
    \centering
    \includegraphics[trim=2.75cm 0.0cm 1.75cm 1.7cm, clip, width=0.245\linewidth]{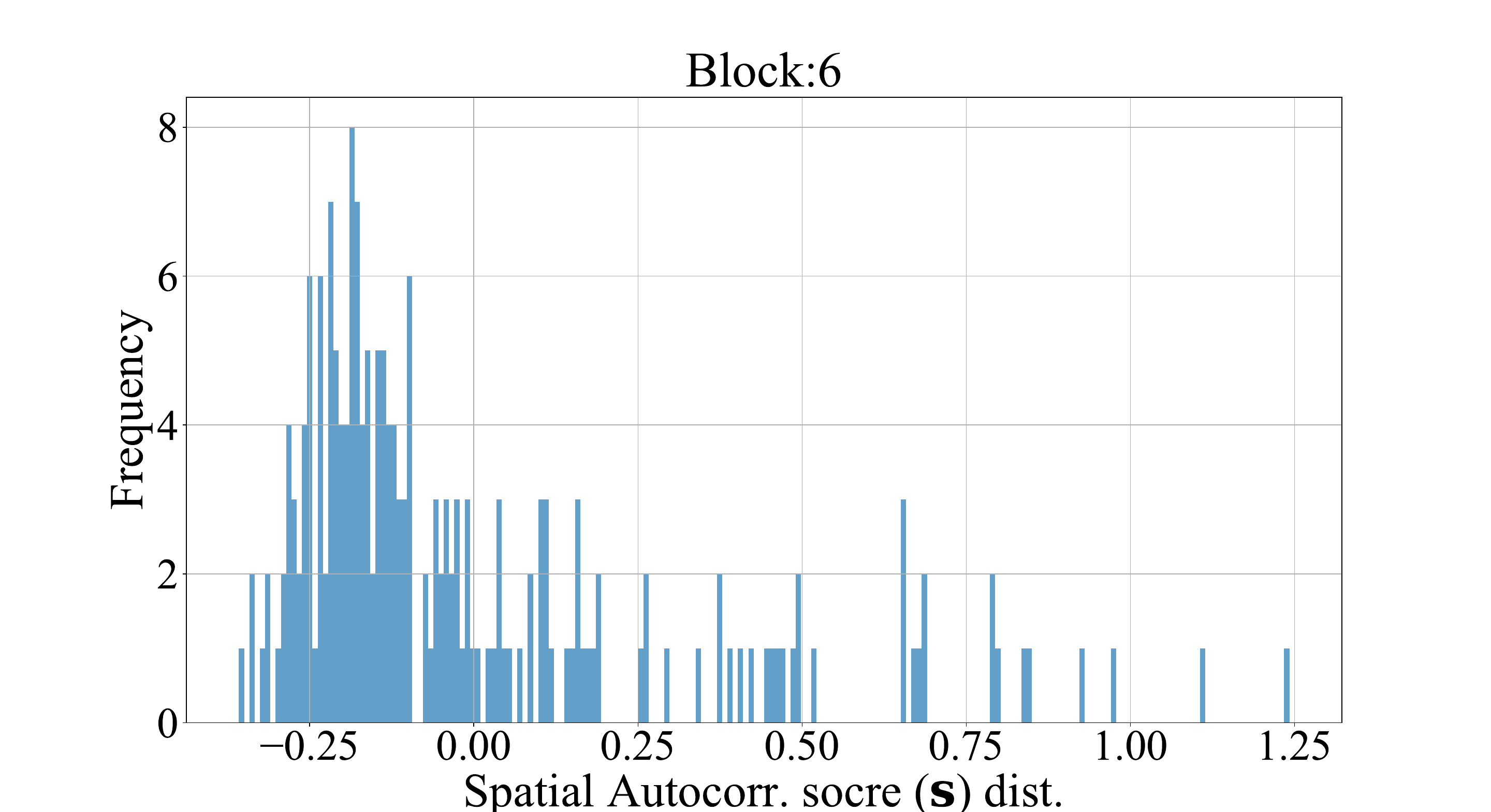}
  \hfill
    \centering
    \includegraphics[trim=2.75cm 0.0cm 1.75cm 1.7cm, clip, width=0.245\linewidth]{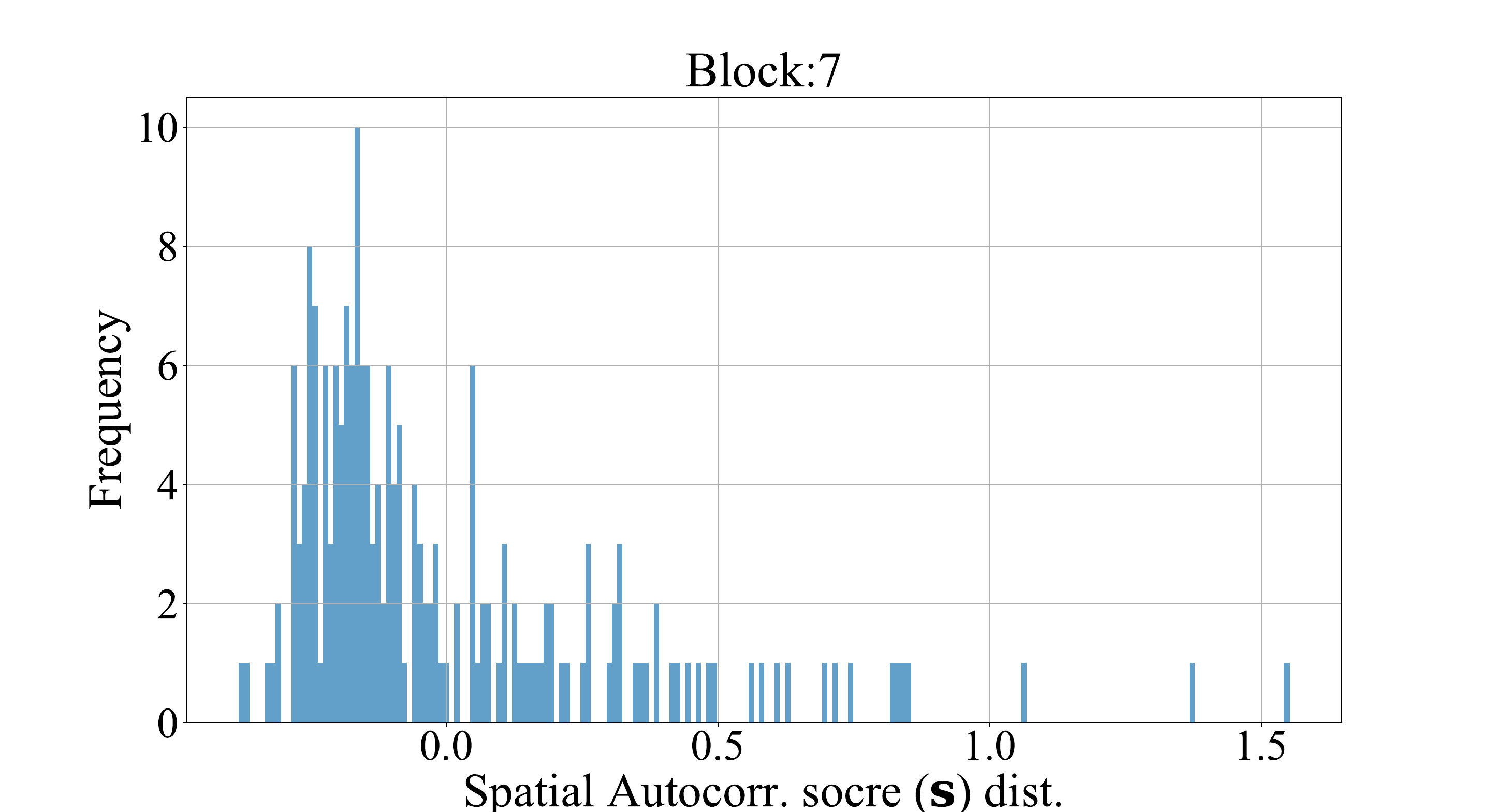}
 \hfill
    \centering
    \includegraphics[trim=2.75cm 0.0cm 1.75cm 1.7cm, clip, width=0.245\linewidth]{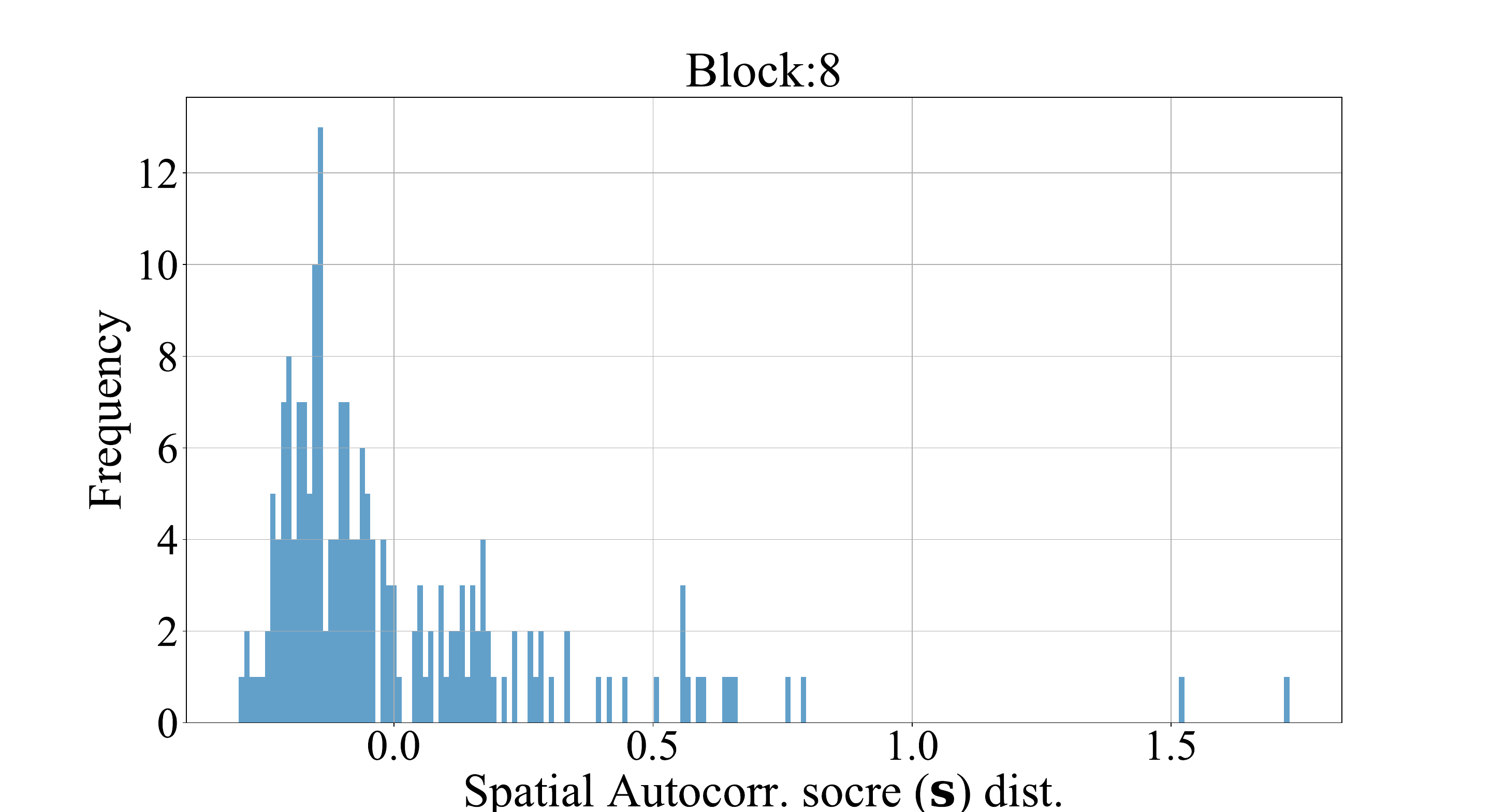} 
  \end{subfigure}
  
\vspace{0.5em} 

  \begin{subfigure}{\textwidth}
    \centering
    \includegraphics[trim=2.75cm 0.0cm 1.75cm 1.7cm, clip, width=0.245\linewidth]{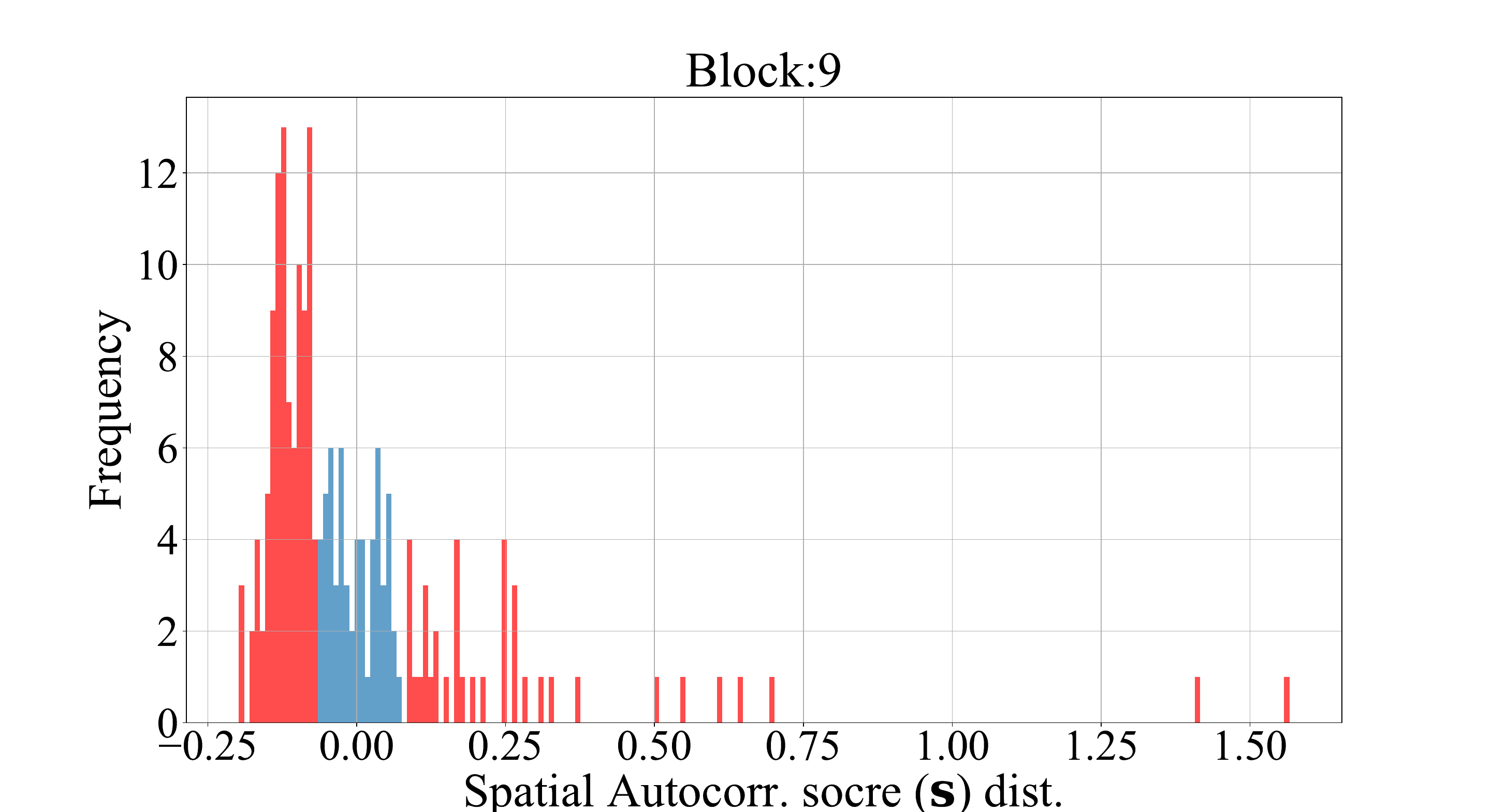}    
  \hfill
    \centering
    \includegraphics[trim=2.75cm 0.0cm 1.75cm 1.7cm, clip, width=0.245\linewidth]{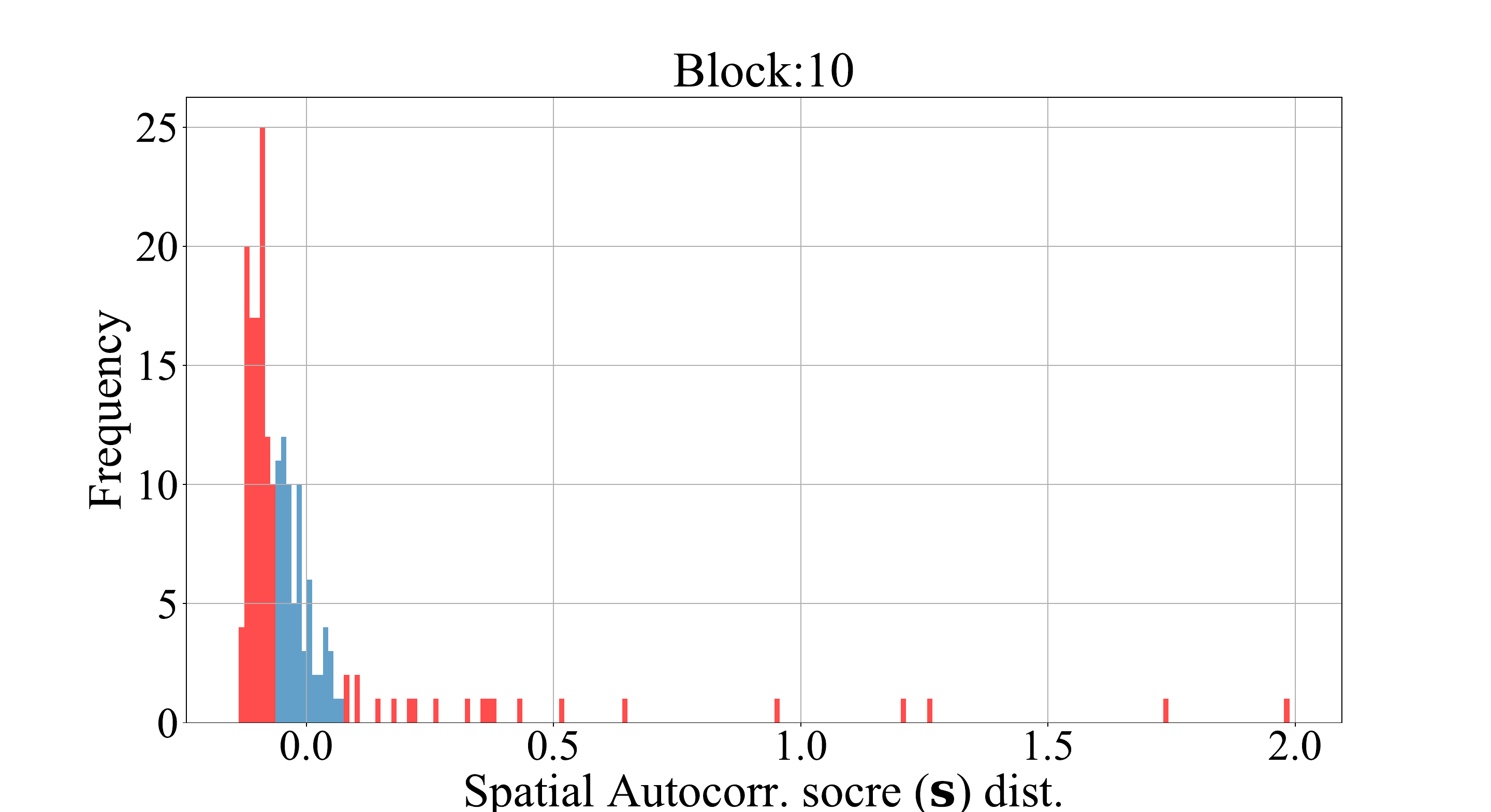}
  \hfill
    \centering
    \includegraphics[trim=2.75cm 0.0cm 1.75cm 1.7cm, clip, width=0.245\linewidth]{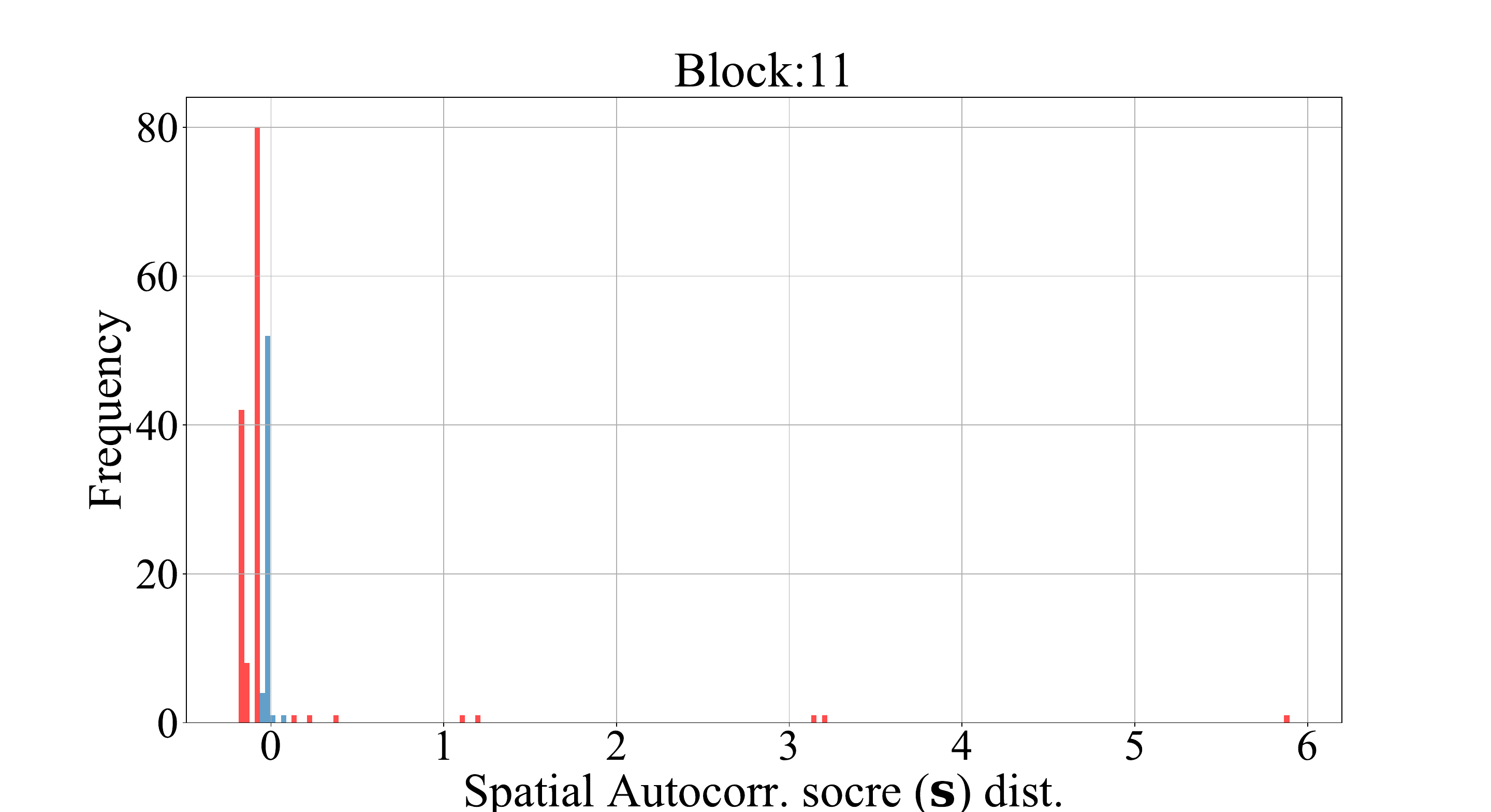}
 \hfill
    \centering
    \includegraphics[trim=2.75cm 0.0cm 1.75cm 1.7cm, clip, width=0.245\linewidth]{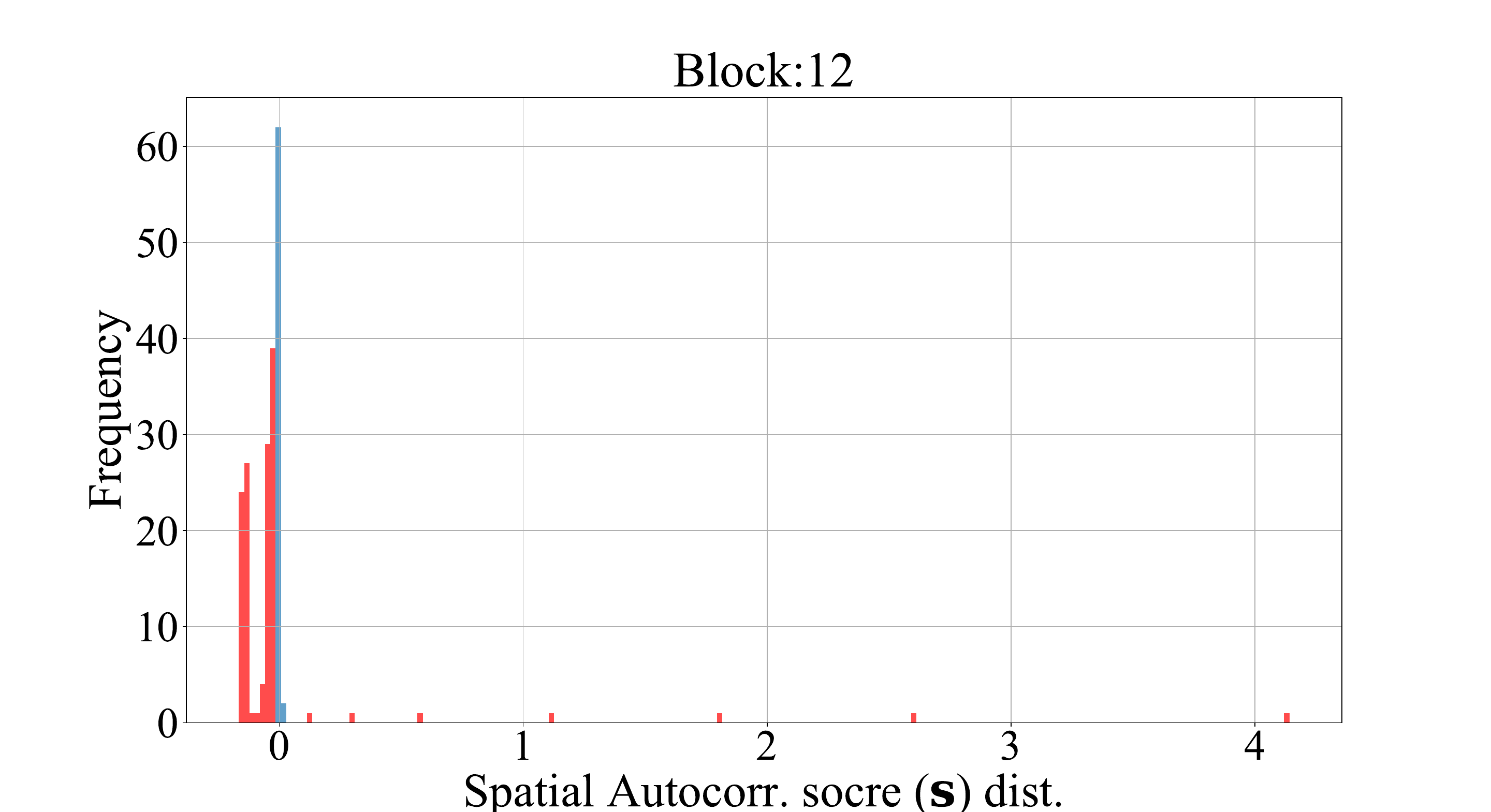} 
  \end{subfigure}
\caption{Visualising the distribution of spatial autocorrelation scores ($\mathbf{s}$) for patches (tokens) generated by various blocks of Deit-Base/16 on the ImageNet-1K validation set. In the last four blocks, tokens with $\mathbf{s}$ scores falling outside of the lower bound ($\mu_{\mathbf{s}} - |\mathbf{\hat{s}}|$) and upper bound ($\mu_{\mathbf{s}} + |\mathbf{\hat{s}}|$) are highlighted in red for the SATA process.}
\label{fig:spt_dist}
  \vspace{1cm} 
\end{figure}

Figure~\ref{fig:spt_dist} shows the distribution of spatial autocorrelation scores ($\mathbf{s}$) for patches (tokens) generated by different blocks of DeiT-Base/16 on the ImageNet-1K validation set. The spatial autocorrelation of tokens decreases through the blocks of the vision transformer, confirming the trends of the upper and lower bands through the vision transformer layers as discussed and demonstrated in Section~\ref{sec:SATA} and Figure~\ref{fig:spt_stage} of the main text. 

\subsection{Robustness on Individual Corruption Type}
\begin{figure}[t]
    \centering
    \includegraphics[trim=1cm 3.5cm 0.5cm 0.5cm, clip, width=\linewidth]{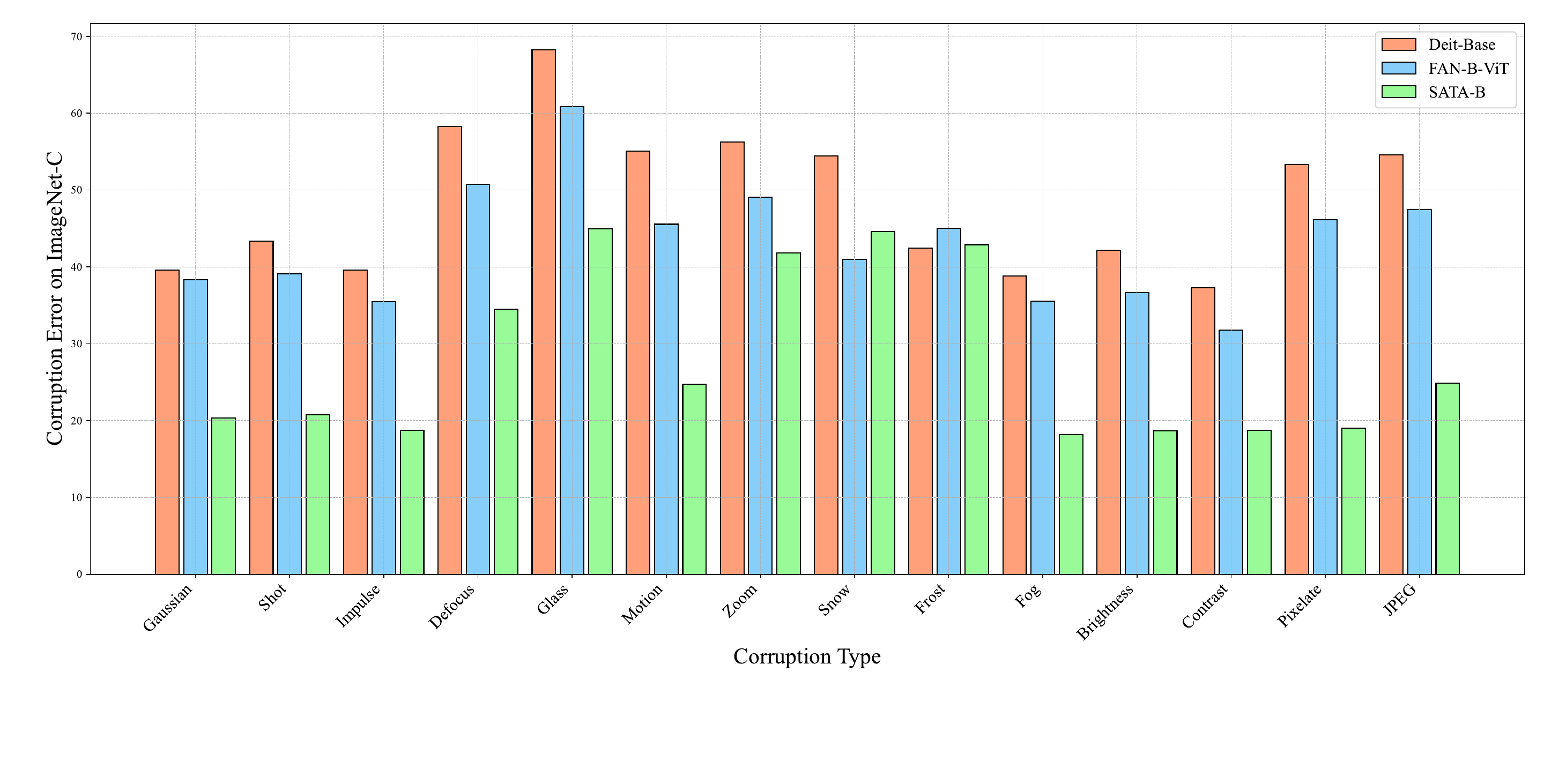}
    \caption{Comparisons of corruption error (the lower, the better) on individual corruption type of ImageNet-C between Deit-Base~\cite{DeiT}, FAN-B-ViT~\cite{FAN} and our SATA-B. Our SATA model significantly outperforms the other baseline models on all of the corruption types.}
    \label{fig:imagenet-c}
      \vspace{1cm} 
\end{figure}
In this experiment, we compare the corruption error on each individual corruption type of ImageNet-C between the baseline DeiT-Base~\cite{DeiT}, FAN-B-ViT~\cite{FAN}, and our SATA-T. As shown in Figure~\ref{fig:imagenet-c}, our SATA model achieves lower corruption errors than the other two models across all corruption types, except for the snow weather corruption. Notably, despite not utilizing any training or patch noise augmentation, the proposed method demonstrates improved robustness and generalizes well to different types of corruption.

\newpage
\subsection{More Visualisation}

\begin{figure}[t]
    \centering
    \includegraphics[trim=0.0cm 0.5cm 0.2cm 0.0cm, clip, width=\linewidth]{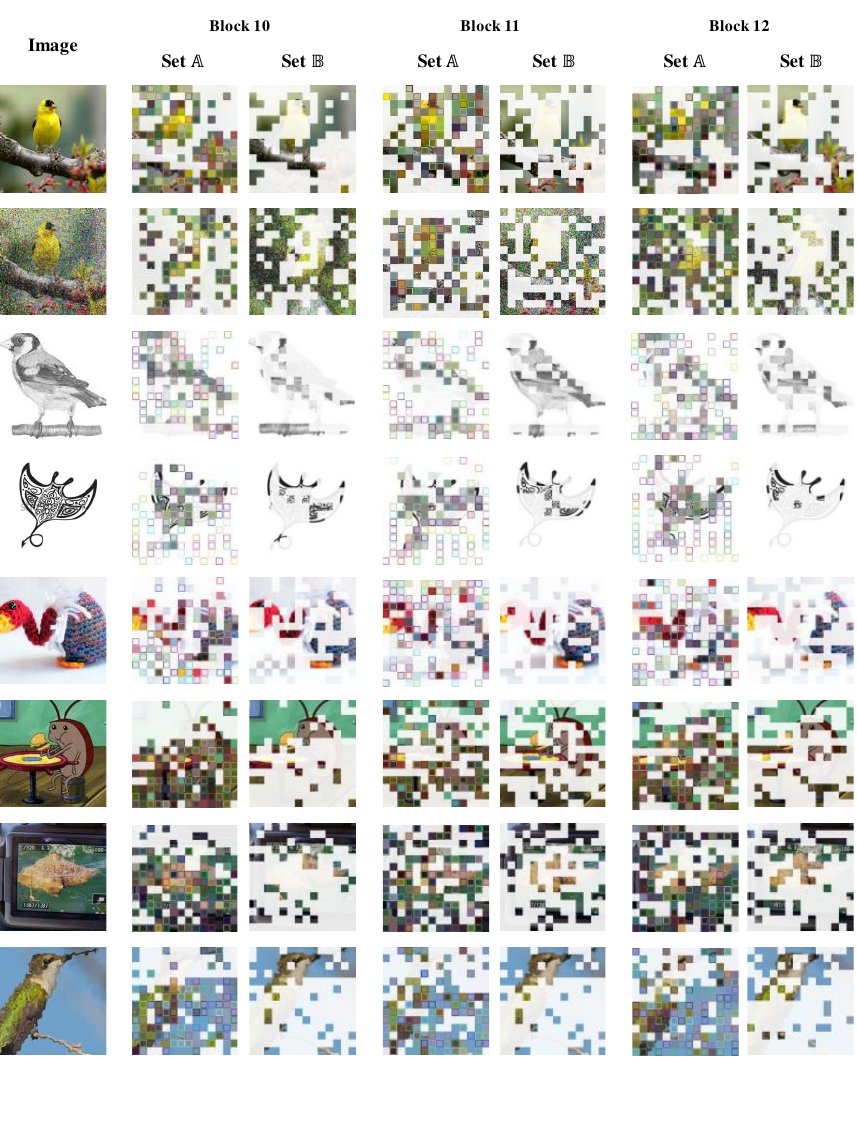}
    \caption{Visualisation of token splitting in Blocks 10 to 12 of SATA-B for images from ImageNet-1K, ImageNet-C, ImageNet-R, ImageNet-A, and ImageNet-SK.}
    \label{fig:setAB_apx}
\end{figure}

\begin{figure}[t]
    \centering
    \includegraphics[trim=0cm 0.5cm 0.cm 0.cm, clip, width=\linewidth]{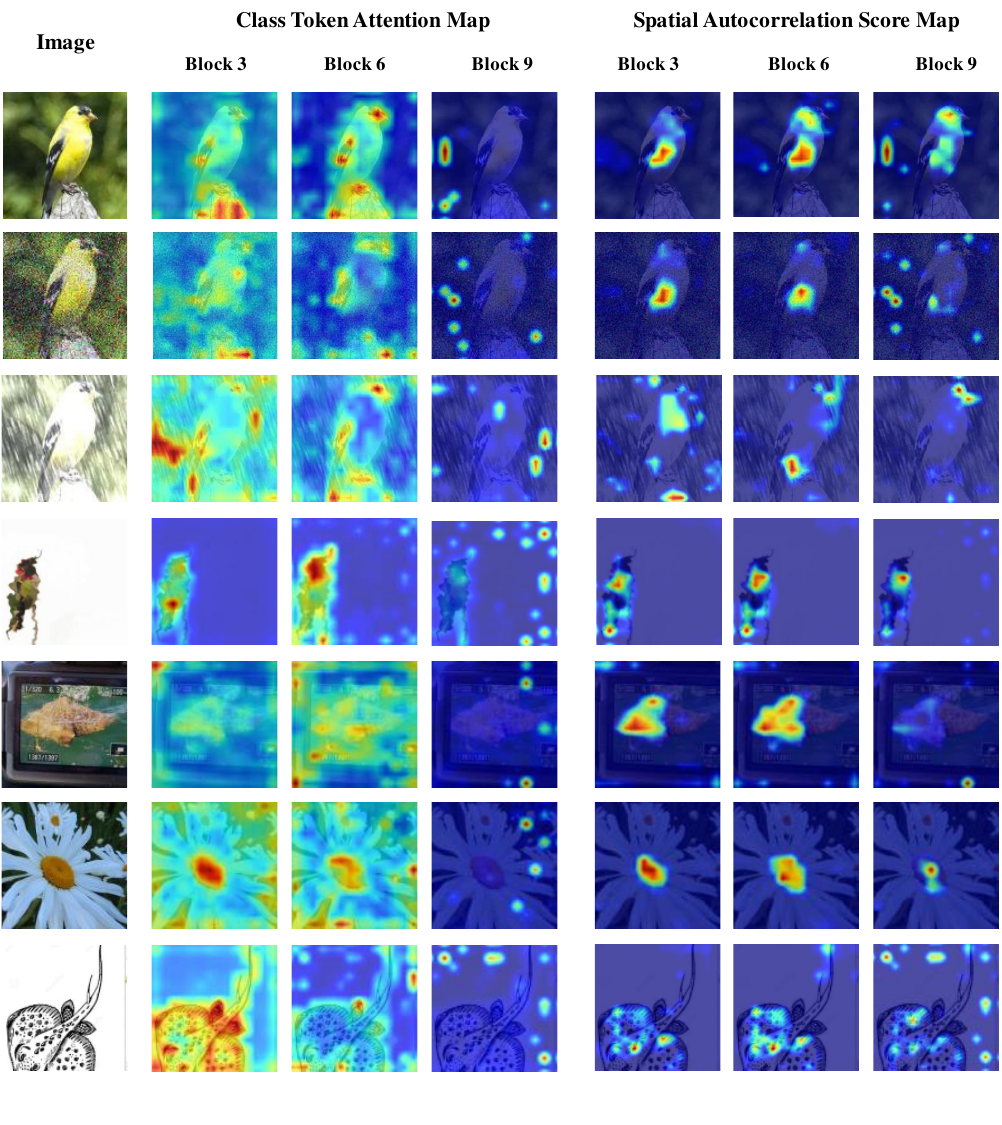}
    \caption{Visual comparison of class token attention maps and spatial autocorrelation score maps across three layers—representing early, middle, and later blocks—of the proposed SATA-B (pre-trained Deit-Base/16+SATA) for different images from ImageNet-1K, ImageNet-C, ImageNet-R, ImageNet-A, and ImageNet-SK.}
    \label{fig:attn_spa_apx}
\end{figure}

To create the visualizations in Figure~\ref{fig:attn_spa_AB}(b) and Figure~\ref{fig:setAB_apx}, we followed the method described in~\cite{ToMe}. We traced each token of Sets $\mathbb{A}$ and $\mathbb{B}$ (described in Eq.\ref{eq:upper-lower}, subsection\ref{sec:token-group}) back to its original input patches. For each token in Set $\mathbb{A}$, we coloured its input patches using the average colour of the tokens it merged with. To distinguish different tokens, we assigned a random border colour to each of the merged tokens.

Moreover, we visualize the comparison of class token attention maps and spatial autocorrelation score maps across three layers—representing early, middle, and later blocks—of the proposed SATA-B (pre-trained Deit-Base/16+SATA) for various images from ImageNet-1K, ImageNet-C, ImageNet-R, ImageNet-A, and ImageNet-SK. As shown in Figure~\ref{fig:attn_spa_apx}, spatial autocorrelation scores exhibit greater consistency across the Transformer layers compared to the corresponding attention scores, suggesting that the use of spatial autocorrelation can provide a more stable and reliable feature representation throughout the network.

\subsection{Implementation}\label{subsec-apx:code}
The following is an implementation of our "Spatial Autocorrelation Token Analysis" (SATA) in PyTorch~\cite{pytorch}. The complete implementation and results are available at https://github.com/nick-nikzad/SATA.

\begin{figure}
    \centering
    \includegraphics[trim=0cm 0.cm 0.cm 0.cm, clip, width=\linewidth]{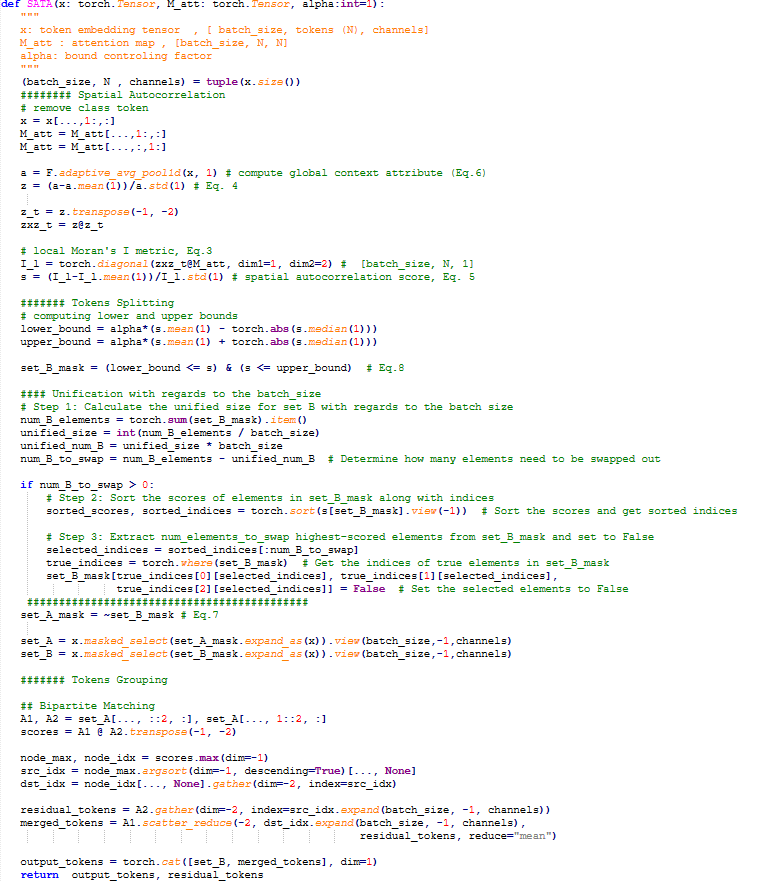}
    \label{fig:implement-code}
\end{figure}

\end{document}